\documentclass[letterpaper]{article}

\usepackage{natbib,alifeconf}  
\usepackage{graphicx}
\usepackage{amsmath}
\usepackage{amsfonts}
\usepackage{comment}
\usepackage{tikz}
\usetikzlibrary{arrows, decorations.markings}
\let\svtikzpicture\tikzpicture
\def\tikzpicture{\noindent\svtikzpicture}
\usetikzlibrary{calc}
\usepackage{array}
\usepackage{hyperref}
\usepackage{supertabular}
\usepackage{nameref}
\usepackage{multicol}   
\usepackage{subcaption}

\def\N{\mathbb{N}}

\makeatletter 
\def\mod@estimate@lineht{%
  \ST@lineht=\arraystretch \baslineskp 
  \ST@stretchht\ST@lineht\advance\ST@stretchht-\baslineskp 
  \ifdim\ST@stretchht<\z@\ST@stretchht\z@\fi 
  \ST@trace\tw@{Average line height: \the\ST@lineht}%
  \ST@trace\tw@{Stretched line height: \the\ST@stretchht}%
} 
\newenvironment{strictsupertabular} 
  {\let\estimate@lineht\mod@estimate@lineht\supertabular} 
  {\endsupertabular} 
\makeatother

\definecolor{myazure}{RGB}{0, 102, 102}

\title{Classification of Discrete Dynamical Systems Based on Transients}
\author{Barbora Hudcová$^{1, 2}$ \and Tomáš Mikolov$^{2}$\\
\mbox{}\\
$^1$Charles University, Prague\\
$^2$Czech Institute of Informatics, Robotics and Cybernetics, CTU, Prague\\
bara.hudcova@gmail.com, tmikolov@gmail.com} 

\tikzstyle{vecArrow} = [thick, decoration={markings,mark=at position
   1 with {\arrow[semithick]{open triangle 60}}},
   double distance=1.4pt, shorten >= 5.5pt,
   preaction = {decorate},
   postaction = {draw,line width=1.4pt, white,shorten >= 4.5pt}]
\tikzstyle{innerWhite} = [semithick, white,line width=1.4pt, shorten >= 4.5pt]

\begin{document}

\maketitle

\begin{abstract}
In order to develop systems capable of artificial evolution, we need to identify which systems can produce complex behavior.
We present a novel classification method applicable to any class of deterministic discrete space and time dynamical systems.
The method is based on classifying the asymptotic behavior of the average computation time in a given system before entering a loop. We were able to identify a critical region of behavior that corresponds to a phase transition from ordered behavior to chaos across various classes of dynamical systems. To show that our approach can be applied to many different computational systems, we demonstrate the results of classifying cellular automata, Turing machines, and random Boolean networks. Further, we use this method to classify 2D cellular automata to automatically find those with interesting, complex dynamics.

We believe that our work can be used to design systems in which complex structures emerge. Also, it can be used to compare various versions of existing attempts to model open-ended evolution (\cite{tierra}, \cite{avida}, \cite{geb}).
\end{abstract}

\section{Introduction}
There are many approaches to searching for systems capable of open-ended evolution. One option is to carefully design a model and observe its dynamics. Iconic examples were designed by \cite{tierra}, \cite{avida}, \cite{geb}, or \citet{chromaria}. However, as we lack any universally accepted formal definition of open-endedness or complexity, there is no formal method of proving the system is indeed ``interesting''. Conversely, lacking definitions of such key terms, it seems extremely difficult to design such models systematically.

Approaching the problem of searching for open-endedness bottom up, we can define a suitable classification of dynamical systems that would help us identify a region of complexity. An ideal classification would be based on a formally defined property, be effectively computable, and help us automatically search for complex systems possibly capable of modeling artificial evolution.

Over the years, many different metrics have been introduced to study systems' dynamics. As an example, cellular automata were studied in terms of their space-time dynamics observations (\cite{wolfram_class}), their space-time compression sizes (\cite{zenil_compression}), via their actions on probability measures (\cite{gutovitz_hier}), the Z-parameter (\cite{wuensche_global}), or the lambda parameter (\cite{langtonAL}). Most of such approaches show that the complex region of systems lies somewhere ``in between'' the ordered and chaotic phase. 

In this paper, we introduce a novel method of classifying complex systems based on estimating their asymptotic average computation time with increasing space size. The key result is that the classification identifies a region of systems that seem to be at a phase transition between ordered and chaotic behavior. Across various classes of discrete systems, we demonstrate that complex systems such as cellular automata computing nontrivial tasks, universal Turing machines, or random Boolean networks at a critical phase belong to this region. Even though we are far from characterizing complexity, we hope this method helps us understand which formally defined properties correlate with it.

\section{Transient Classification: A General Method} \label{general_method}
We first introduce the basic principle of the classification based on transients, which can be applied to any deterministic discrete space and time dynamical system (DDDS). In subsequent sections, we describe the results of the classification applied to cellular automata, Turing machines, and random Boolean networks to demonstrate its use across different classes of discrete dynamical systems.

\subsubsection{Basic Notions}
Let us consider a generic deterministic discrete system $D$ operating on finite space, characterized by a tuple $D =(S, F)$ where $S$ is a finite set of configurations and $F: S \rightarrow S$ is a global transition function governing the dynamics of the system. We define the \textit{trajectory of a configuration }$u \in S$  as the sequence
$$(u, F(u), F^2(u), \ldots).$$
As $S$ is finite, every trajectory eventually becomes periodic. We call the preperiod of this sequence the \textit{transient of initial configuration $u$} and denote its length by $t_u$. More formally, we define $t_u$ to be the smallest positive integer $i$, for which there exist $j \in \N$, $j > i$, such that $F^i(u) = F^j(u)$. The periodic part of the sequence is called an \textit{attractor}. The \textit{phase-space} of $D=(S, F)$ is an oriented graph with vertices $V = S$ and edges $E = \{(u, F(u)), u \in S\}$. Such a graph is composed of components, each containing one attractor and multiple transient paths leading to the attractor. The phase-space completely characterizes the dynamics of the system. However, it is infeasible to describe when the configuration space $S$ is large. Given a DDDS $D$,  we will focus on studying its \textit{average transient length}
$$T(D) = \frac{1}{|S|} \sum_{u \in S} t_u.$$
We describe the method of estimating a system's average transient length together with the error analysis in section \nameref{stats}.

\subsubsection{The Main Principle}
Suppose we have a sequence of DDDSs $$D_1 = (S_1, F_1), D_2 = (S_2, F_2), D_3 = (S_3, F_3), \, \ldots$$
operating on configuration spaces of growing size. That is, $F_i: S_i \rightarrow S_i$ and $|S_i| < |S_{i+1}|$ for each $i$. For instance, the sequence can be given by a cellular automaton with a fixed local rule, operating on a finite cyclic grid of growing size.
 
Our goal is to estimate the asymptotic growth of the systems' average transient lengths, as shown in Figure \ref{main_method_diagram}. 

\begin{figure}[h!] 
\centering
\begin{tikzpicture}
\node[thick] at (-1, 3) {\textbf{Discrete system}}; \node[thick] at (3, 3) {\textbf{Average transient}}; 
																			\node at (3, 2.6) {\textbf{length}};
\node at (-1, 2) {$D_1 = (S_1, F_1)$}; \node at (3, 2) {$T(D_1)$}; 
\node at (-1, 1.4) {$D_2 = (S_2, F_2)$}; \node at (3, 1.4) {$T(D_2)$}; 
\node at (-1, 0.8) {$D_3 = (S_3, F_3)$}; \node at (3, 0.8) {$T(D_3)$}; 
\node at (-1, 0.2) {$D_4 = (S_4, F_4)$}; \node at (3, 0.2) {$T(D_4)$}; 
\node at (-1, -.4) {$\vdots$}; \node at (2.7, -.4) {$\vdots$}; 
\path[->, very thick] (6, 2.1) edge node[sloped, anchor=center, above] {\textbf{asymptotic growth}} (6, -.4);
\end{tikzpicture}
\caption{   Diagram depicting the asymptotic growth of average transient lengths of a sequence of discrete systems.}
\label{main_method_diagram}
\end{figure}

\begin{figure}[h!]
\centering
\scriptsize
\begin{tikzpicture}[thick, every node/.style={inner sep=0,outer sep=0}]

\draw[rounded corners, fill=myazure!5] (-.7, -0.8) rectangle (2., .8) {};
\draw[rounded corners, fill=myazure!20] (2.4, -0.8) rectangle (5.1, .8) {};
\draw[rounded corners, fill=myazure!50] (5.5, -0.8) rectangle (6.9, .8) {};

\node[] at (0.7, 1) {\textbf{Ordered Phase}};
\node at (0, 0) {\includegraphics[width=0.16\linewidth]{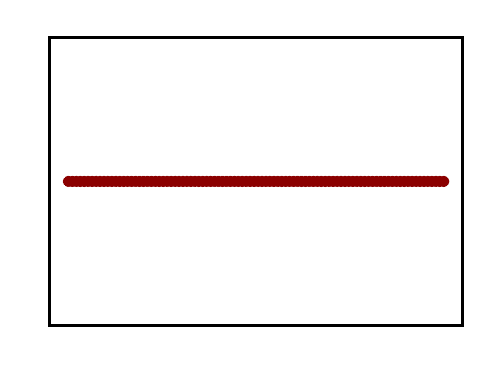}};
\node[] at (0, .6) {Bounded};

\node at (1.3, 0) {\includegraphics[width=0.16\linewidth]{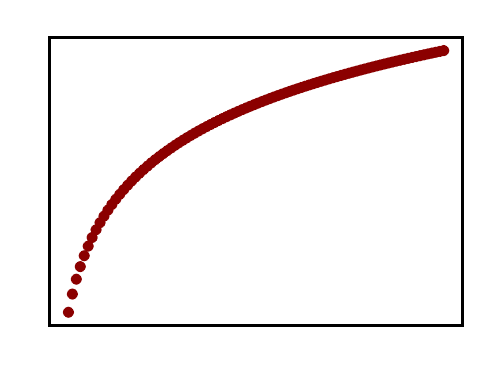}};
\node[] at (1.3, .58) {Log};

\node[] at (3.8, 1) {\textbf{Phase Transition}};
\node at (3.1, 0) {\includegraphics[width=0.16\linewidth]{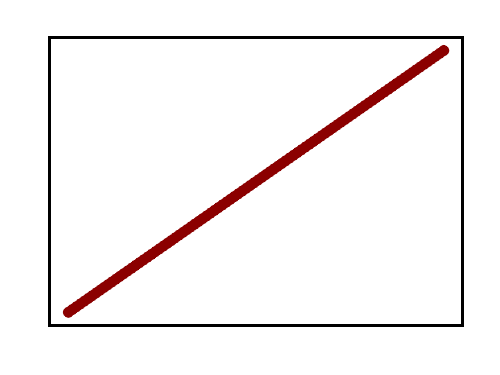}};
\node[] at (3.1, .6) {Lin};

\node at (4.4, 0) {\includegraphics[width=0.16\linewidth]{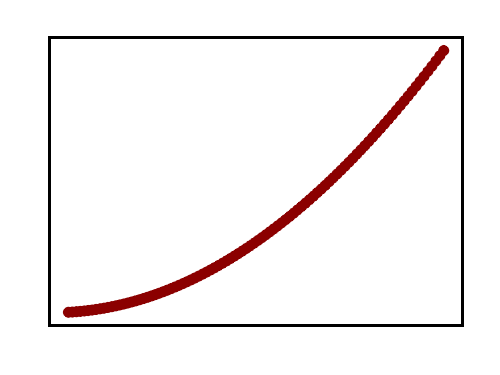}};
\node[] at (4.4, .58) {Poly};

\node[] at (6.2, 1) {\textbf{Chaotic Phase}};
\node at (6.2, 0) {\includegraphics[width=0.16\linewidth]{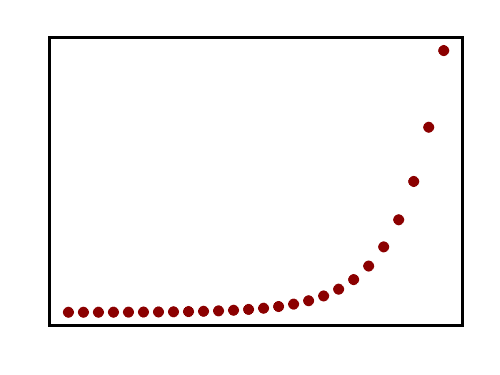}};
\node[] at (6.2, .6) {Exp};

\draw (0.7, -.95) edge (0.7, -1.05);
\node at (3.45, -1.2) {Chaos Increasing};

\draw[vecArrow] (0.7, -1) to (6.2, -1);

\end{tikzpicture}
\caption{General trend of the transient classification results.}
\label{trans_trend_diagram}
\end{figure}
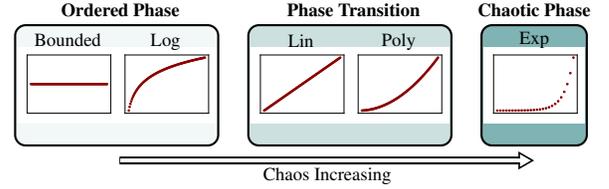

In practice, we generate a finite part of the sequence $(|S_i|, T(D_i))_{i=1}^B$ where $B$ is an upper bound imposed by our computational limitations and examine different regression fits of the data. Specifically, we evaluate the fit to constant, logarithmic, linear, polynomial, and exponential functions. We pick the best fit with respect to the $R^2$ score and obtain the classes: Bounded, Log, Lin, Poly, and Exp. If the score of the fit to all such functions is low (i.e., $R^2<85\%$), we say the system is Unclassified. Surprisingly, we found a very good fit to one of the classes with $R^2 > 90\%$ for most DDDSs we examined. The trend we have observed, which seems to hold across various families of DDDSs, is shown in Figure \ref{trans_trend_diagram}.  We describe it in more detail for each family in the subsequent sections.

We do not claim our method determines the true asymptotic behavior of a system; it is merely a possible interpretation of the method. For some systems, the transient growth might correspond to more complicated functions but we have deliberately chosen the classes to be quite robust and coarse to have clearer boundaries between them. The uncertainty of the true asymptotic growth is especially relevant for the Lin and Poly Classes which identify the critical phase transition region. Such systems might turn out to be logarithmic or exponential, and it might be the case that we have not detected this due to our limited data. However, in such a case, such systems would exhibit significantly slower convergence to their asymptotic behavior than systems in other classes which is a typical property of a system at a phase transition.

\subsubsection{Computational Interpretation}
In non-classical models of computation (\cite{natural_comp}), the process of traversing a discrete system's transients can be perceived as the process of self-organization, in which information can be aggregated in an irreversible manner. The attractors are then viewed as memory storage units, from which the information about the output can be extracted. For cellular automata (CAs), this is explored in \cite{kaneko}. The average transient growth then corresponds to the average computation time of the system\footnotemark.
\footnotetext{Here, the computation time is understood in the abstract sense; as the number of iterations of the transition function.} Therefore, we can interpret our goal as trying to estimate systems' \textbf{asymptotic average computation time}. DDDSs with bounded transient lengths can only perform trivial computation. On the other hand, DDDSs with exponential transient growth can be interpreted as inefficient computation models.

In the context of artificial evolution, we can view the global transition rule of a DDDS as the physical rule of the system, whereas the initial configuration as the particular ``setting of the universe'', which is then subject to evolution. If we are interested in finding DDDSs capable of complex behavior automatically, it would be beneficial for us if such behavior occurred on average, rather than having to select the initial configurations carefully from some narrow region. The probability of finding such special initial configurations would be extremely low as the configuration space tends to be very large. This motivates our study of the growth of average transient lengths rather than the maximum ones.

\subsubsection{Average Transients: Error Estimate} \label{stats}
Let us fix a DDDS $D=(S, F)$ operating on a large configuration space, e.g., $|S| \gg 2^{100}$. In such case, computing the average transient length $\mu$ is infeasible. Thus, we uniformly randomly sample initial configurations $u_1, u_2, \ldots, u_m$ and estimate $\mu$ by $\frac{1}{m} \sum_{i=1}^m t_{u_i}$. It remains to estimate the number of samples $m$ so that the error $|\frac{1}{m} \sum_{i=1}^m t_{u_i} - \mu|$ is reasonably small. 

More formally, for $D=(S, F)$, let $(S, P)$ be a discrete probability space where $S$ is the set of all configurations and $P$ is a uniform distribution. Let $ X: S \rightarrow \N$ be a random variable, which sends each $u$ to its transient length $t_u$. This gives rise to a probability distribution of transient lengths on $\mathbb{N}$ with mean $E(X)$ and variance $var(X)$. It can be easily shown that $E(X) = \mu$. Our goal is to obtain a good estimate of $E(X)$ by the Monte Carlo method (\cite{mcbook}).

Let $(X_1, X_2, \ldots, X_m)$ be a random sample of iid random variables, $X_i \,{\buildrel d \over =}\, X$ for all $i$. Let $\mu^{(m)} = \frac{1}{m} \sum_{i=1}^m X_i$ be the sample mean and $\sigma^{(m)} = \sqrt{\frac{1}{m-1} \sum_{i=1}^m (X_i - \mu^{(m)} )^2}$ the sample standard deviation. As $X$ is a mapping from a finite set, $var(X) < \infty$, and thus we have by the Central limit theorem the convergence to a normal distribution. The interval
$$\Big ( \mu^{(m)} - u_{1 - \frac{\alpha}{2}} \frac{\sigma^{(m)} }{\sqrt{m}}, \mu^{(m)} + u_{1 - \frac{\alpha}{2}} \frac{\sigma^{(m)} }{\sqrt{m}} \Big )$$
where $u_{\beta}$ is the $\beta$ quantile of the normalized normal
 distribution, covers $\mu$ for $m$ large with probability approximately $1-  \alpha$. We will take $\alpha = 0.05$. Hence, with probability approximately $95\%$ $$|\mu - \mu^{(m)}| < u_{0.975} \frac{\sigma^{(m)}}{\sqrt{m}}.$$
 
From the nature of our data, both the values $E(X) = \mu$ and $var(X)$ tend to grow with increasing size of the configuration space. Therefore, to employ a general method of estimating the number of samples, we normalize the error by the sample mean and consider $\frac{|\mu - \mu^{(m)}|}{\mu^{(m)}}$. 
 Therefore for $m$ sufficiently large such that 
 \begin{align} \label{sample_cond}
 u_{0.975}  \frac{\sigma^{(m)}}{\sqrt{m}  \mu^{(m)}} < \epsilon 
 \end{align}
 we have that $\mu^{(m)}$ differs from $\mu$ by at most $\epsilon \cdot 100\%$ with probability approximately $95\%$. 

In practice, we put $\epsilon = 0.1$ and produce the observations in batches of size 20 until condition (\ref{sample_cond}) is met (for most elementary CA this was satisfied typically after 400 data points were sampled). We approximate the uniform random sampling of initial configurations using Python's numpy.random library. 
\hyphenation{ap-proximated}

\section{Cellular Automata}
\subsection{Introducing Cellular Automata}
Informally, a cellular automaton (CA) can be perceived as a $k$-dimensional grid consisting of identical finite state automata. They are all updated synchronously in discrete time steps based on an identical local update function depending only on the states of automata in their local neighborhood. A formal definition can be found in \cite{kari_survey}.

CA were first studied as models of self-replicating structures  (\cite{neuman}, \cite{LANGTONselfrep}, \cite{REGGIAselfrep}). Subsequently, they were examined as dynamical systems (\cite{hedlund}, \cite{vichniac}, \cite{gutowitz_local}), or as models of computation (\cite{toffoli}, \cite{mitchell_overview}). Being so simple to simulate, yet capable of complex behavior and emergent phenomena (\cite{crutchfield}, \cite{emergence1}), CA provide a convenient tool to examine the key, yet undefined notions of complexity and emergence.

\subsubsection{Basic Notions}
We study the simple class of \textit{elementary cellular automata} (ECAs), which are one-dimensional CAs with two states $\{0, 1 \}$ and neighborhood of size 3. Each ECA is given by a local transition function $f: \{0, 1 \}^3 \rightarrow \{0, 1 \}$. Hence, there are only 256 of them. The size of this CA class is the reason to make it our first case of study. One can simply explore it by studying the dynamics of every single ECA.

We identify each local rule $f$ determining an ECA with the \textit{Wolfram number} of the ECA defined as:
$$2^0 f(0, 0, 0) + 2^1 f(0, 0, 1) + 2^2 f(0, 1, 0) + \ldots + 2^7 f(1, 1, 1).$$
We will refer to each ECA as a ``rule $k$'' where $k$ is the corresponding Wolfram number of its underlying local rule.

We will consider the CA to operate on finite grids with periodic boundary conditions. Hence, given a local rule $f$ and a grid size $n$, we obtain a configuration space $\{0, 1 \}^n$ and a global update function $F: \{ 0, 1\}^n \rightarrow \{0, 1 \}^n$.

Let $(\{0, 1 \}^n, F)$ be an ECA operating on a grid of size $n$ and $(u, F(u), F^2(u), \ldots)$ a trajectory of a configuration $u \in \{0, 1 \}^n$.
The space-time diagram of such a simulation is obtained by plotting the configurations as horizontal rows of black and white squares (corresponding to states 1 and 0) with a vertical axis determining the time, which is progressing downwards.

We note that properties of CA phase-spaces were examined among others by \cite{wuensche_global}. Precisely for this purpose, a software was designed by \cite{ddlab}.

\hyphenation{ap-proximated}

\subsection{History of CA Classifications} 
An ideal classification would be based on a rigorously defined and efficiently measurable property, identifying a region of systems with interesting behavior.
In this section, we describe three qualitatively different classifications of ECAs, and subsequently, we will compare our results to them.

\hyphenation{ap-proximated}

\subsubsection{Wolfram's Classification}
\begin{figure}[h!]
    \centering
    \begin{tikzpicture}[thick, every node/.style={inner sep=0,outer sep=0}]
  \node at (3.4, 0) {
     \includegraphics[width=0.43\linewidth]{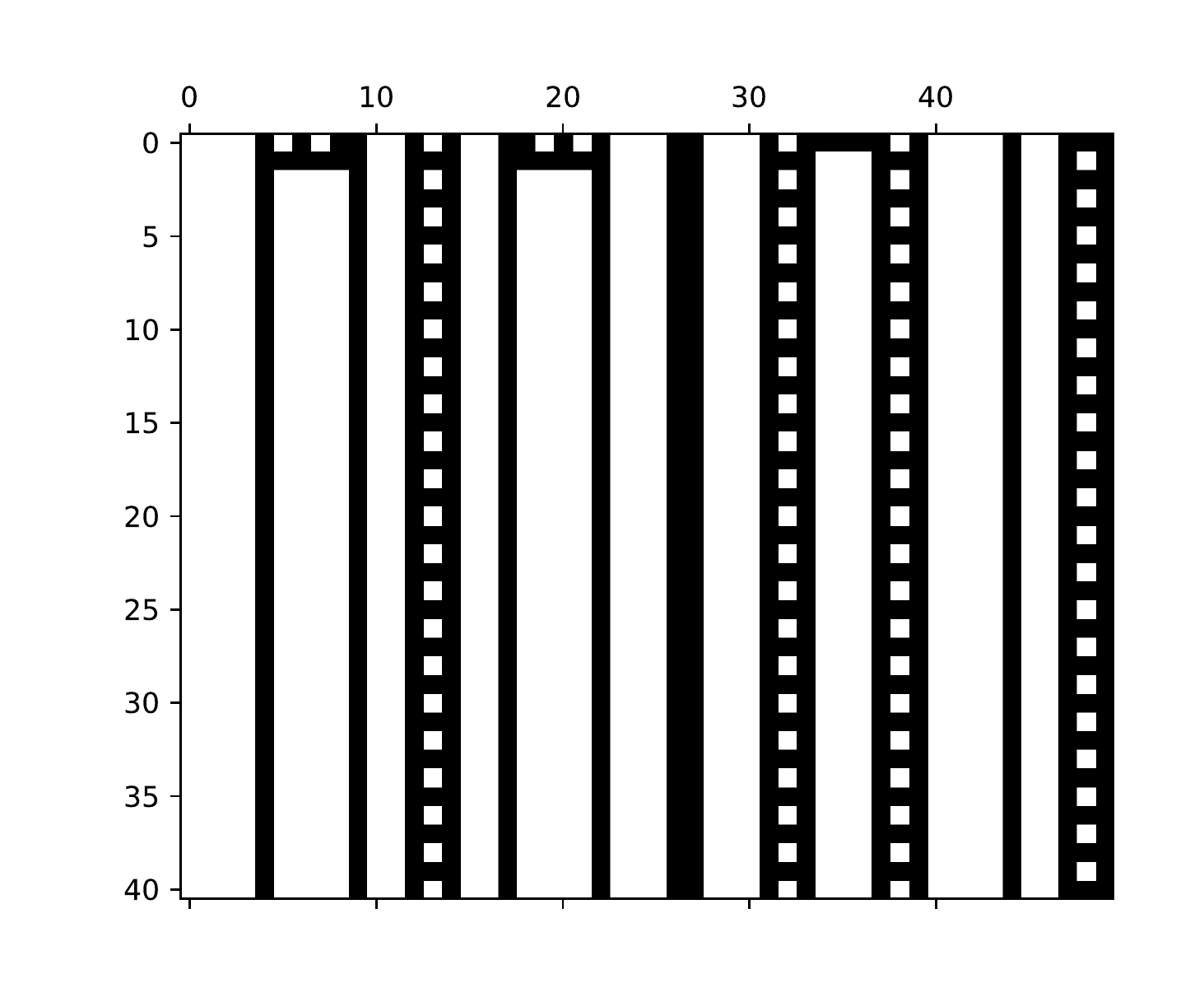}
  };
  \node at (0, 0) {
     \includegraphics[width=0.43\linewidth]{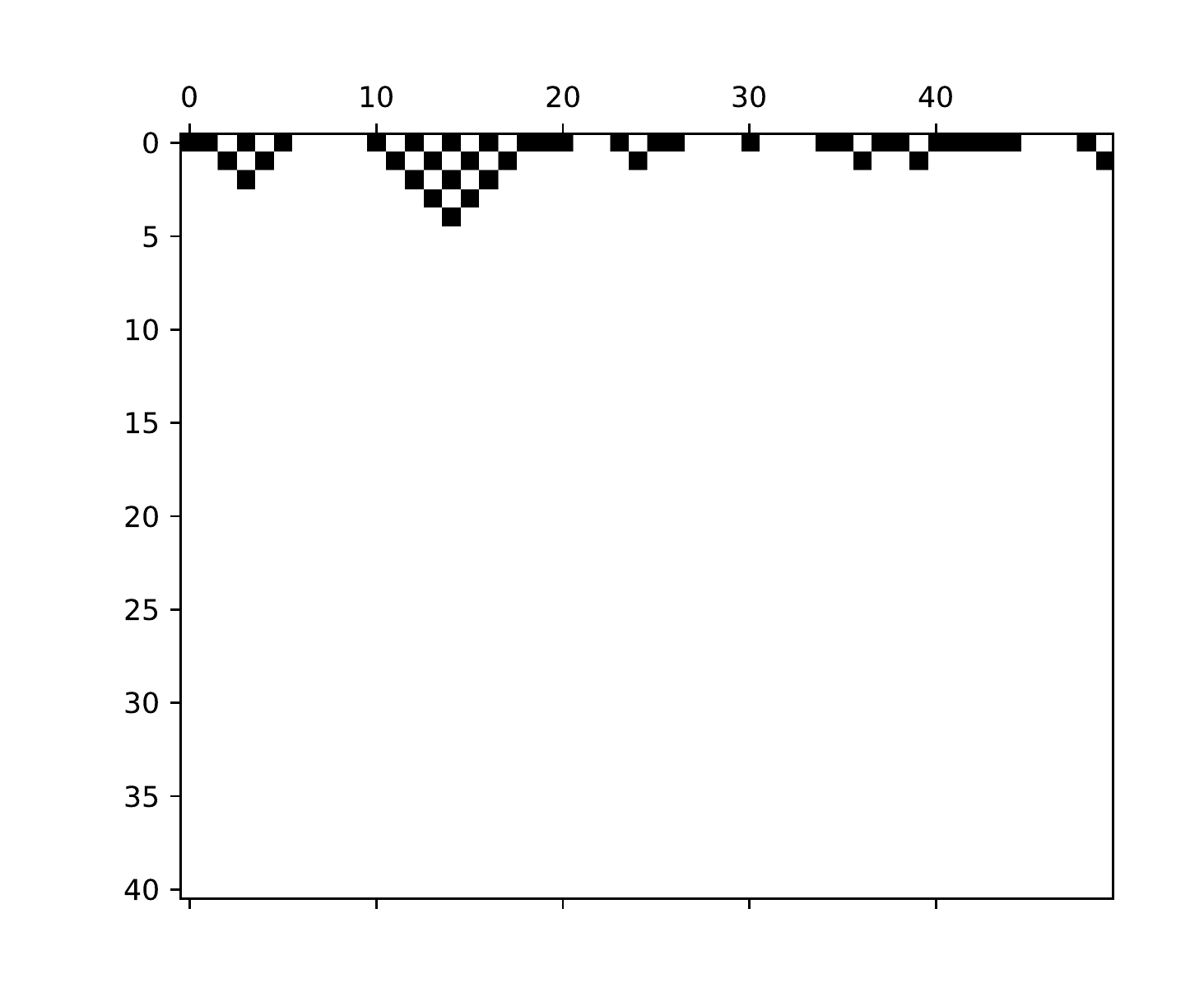}
  };
  \node at (3.4, -2.8) {
     \includegraphics[width=0.43\linewidth]{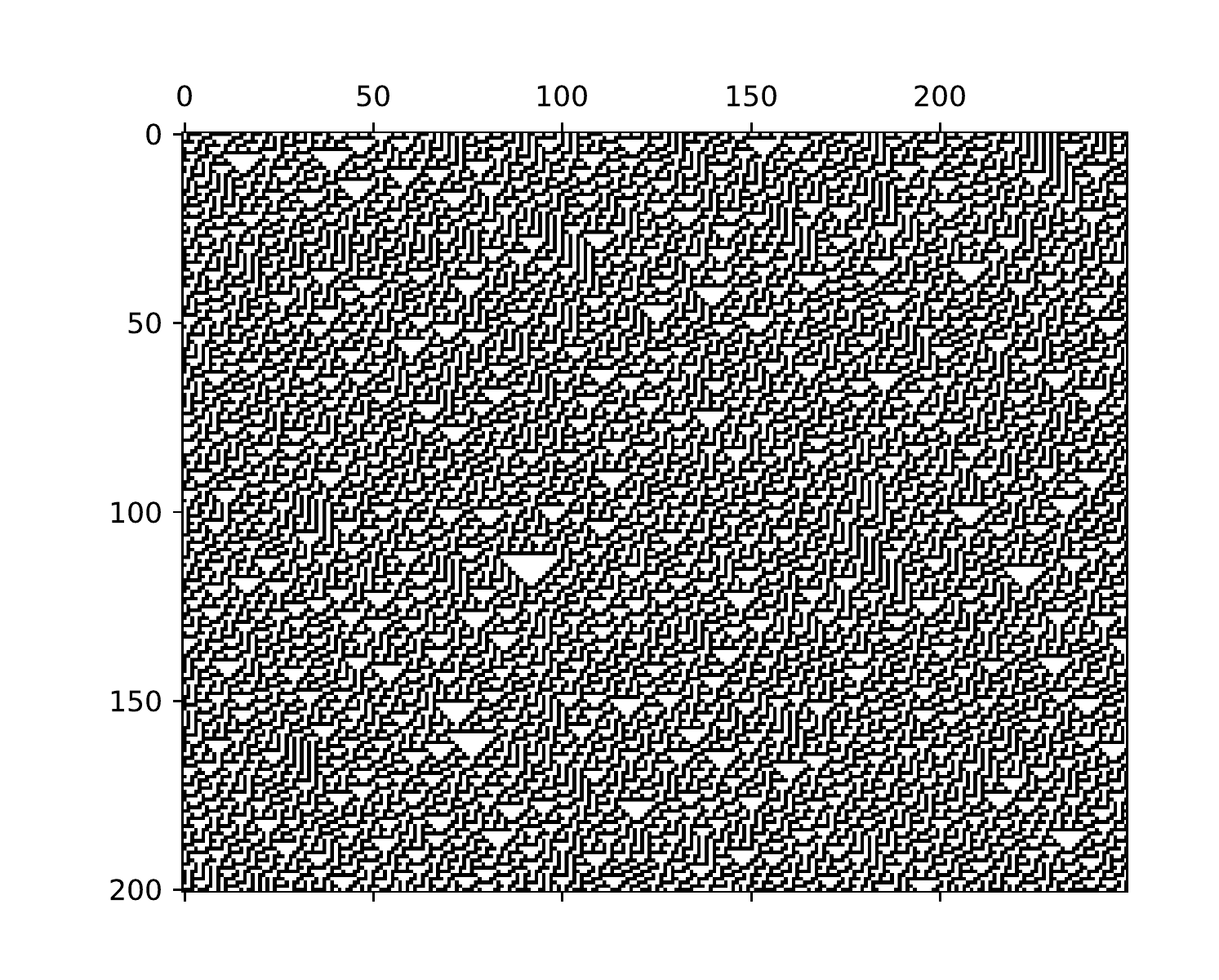}
  };
  \node at (0, -2.8) {
     \includegraphics[width=0.43\linewidth]{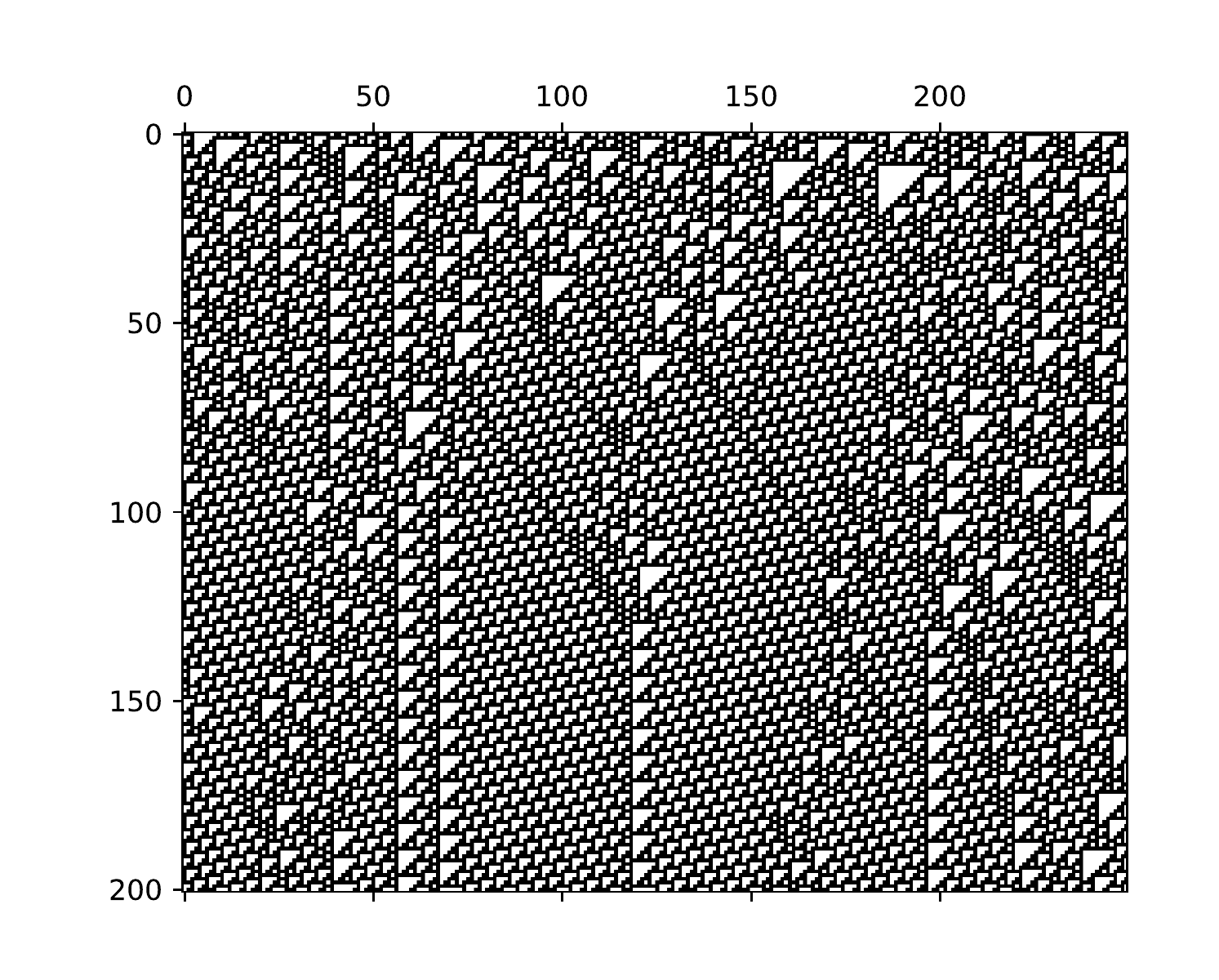}
  };
\end{tikzpicture}
    \caption{Space-time diagrams of rules from each Wolfram's class. Class 1 rule 32 is on top left, Class 2 rule 108 on top right, Class 4 rule 110 on the bottom left, and  Class 3 rule 30 on the bottom right.}
    \label{wolf_classes}
\end{figure}

\begin{figure}[h!]
    \centering
    \begin{tikzpicture}[thick, every node/.style={inner sep=0,outer sep=0}]
  \node at (0, 0) {
     \includegraphics[width=0.5\linewidth]{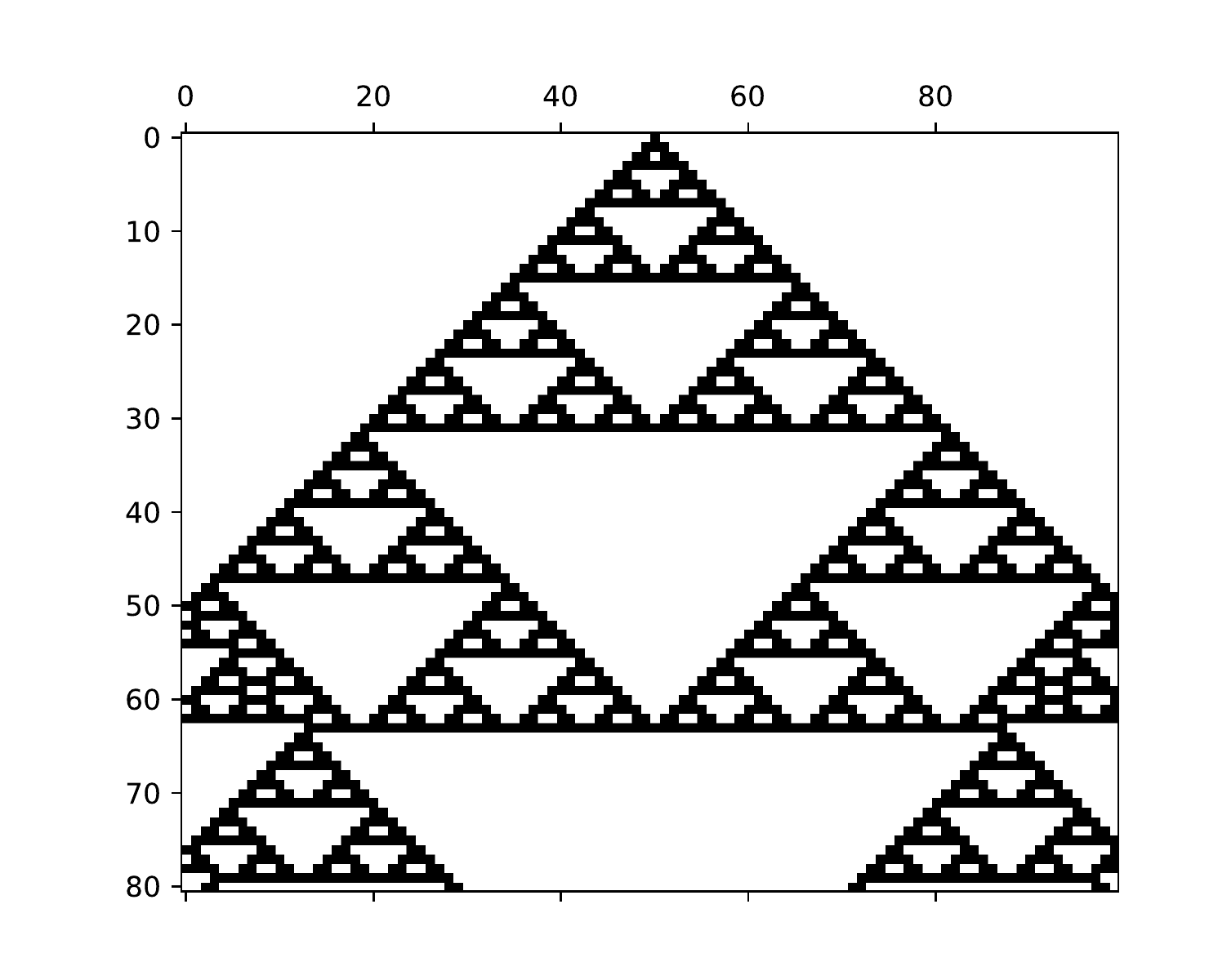}
  };
  \node at (4, 0) {
     \includegraphics[width=0.5\linewidth]{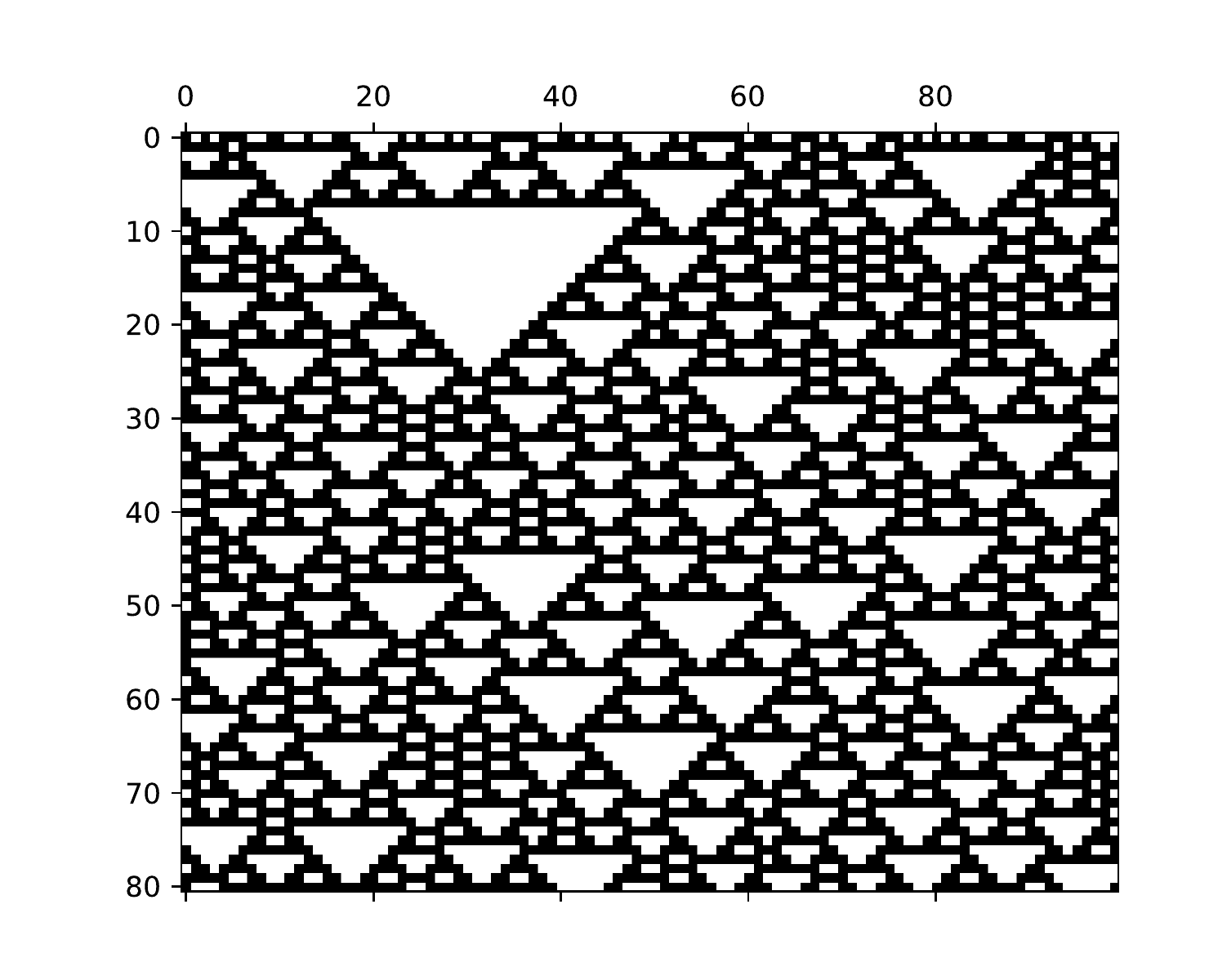}
  };
\end{tikzpicture}
\caption{On the left, rule 126 is simulated with an initial condition consisting of a single 1 bit padded with 0's. On the right, the same rule is simulated with a random initial configuration.}
\label{sensitiveCA}
\end{figure}

The most intuitive and simple approach to examining the dynamics of CAs is to observe their space-time diagrams. This method was particularly proclaimed by \cite{newscience}. Therein, he established an informal classification of CA dynamics based on such diagrams. \cite{newscience} distinguishes the following classes, which are shown in Figure \ref{wolf_classes}.
\begin{align*}
\text{Class 1 } \ldots &\text{ quickly resolves to a homogenous state}\\
\text{Class 2 } \ldots &\text{ exhibits simple periodic behavior}\\
\text{Class 3 } \ldots &\text{ exhibits chaotic or random behavior}\\
\text{Class 4 } \ldots &\text{ produces localized structures that}\\
& \text{ interact with each other in complicated ways}
\end{align*}
The main issue is that we have no formal method of classifying CAs in this way. In fact, this problem is in general undecidable (\cite{culik}). Moreover, the behavior of some CAs can vary with different initial configurations. An example being rule 126 which oscillates between Class 2 and Class 3 behavior, as shown in Figure \ref{sensitiveCA}. The transient classification we present in this paper deals with both these issues. 

\hyphenation{ap-proximated}

\subsubsection{Zenil's Classification}
\begin{figure}[h!]
    \centering
    \begin{tikzpicture}[thick, every node/.style={inner sep=0,outer sep=0}]
  \node at (1.5, 0) {
    \includegraphics[width=0.26\textwidth]{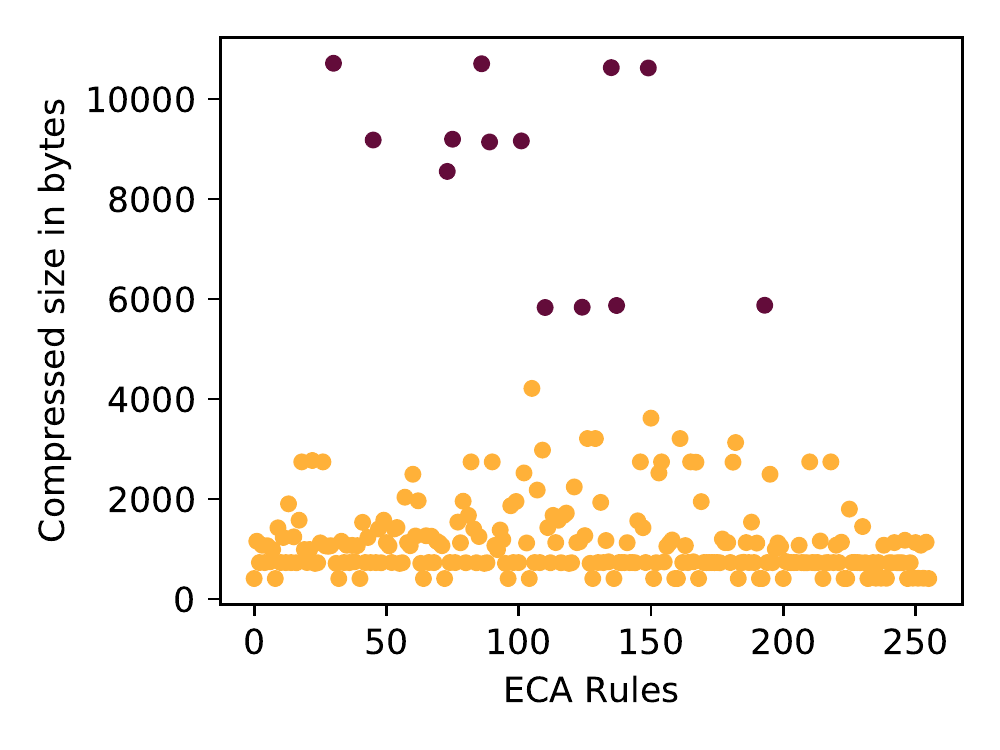}};
\end{tikzpicture}
\caption{Reproduction of Zenil's results (\cite{zenil_compression}). The purple cluster corresponds to the interesting Class 3 and 4 rules, the yellow cluster to the rest.}
\label{zenil_reproduction}
\end{figure}

In the first part of his paper, \cite{zenil_compression} studied the compression size of the space-time diagrams of each ECA simulated for a fixed amount of steps. For the classification, he examines the simulations from a particular initial configuration (a single one surrounded by zeros). Using a clustering technique, he obtained two classes distinguishing between Wolfram's simple classes 1 and 2 and complex classes 3 and 4. We show our reproduction of Zenil's results in Figure \ref{zenil_reproduction}.

\begin{figure}[h!]
    \centering
    \begin{tikzpicture}[thick, every node/.style={inner sep=0,outer sep=0}]

  \node at (0, 0) {
     \includegraphics[width=0.45\linewidth]{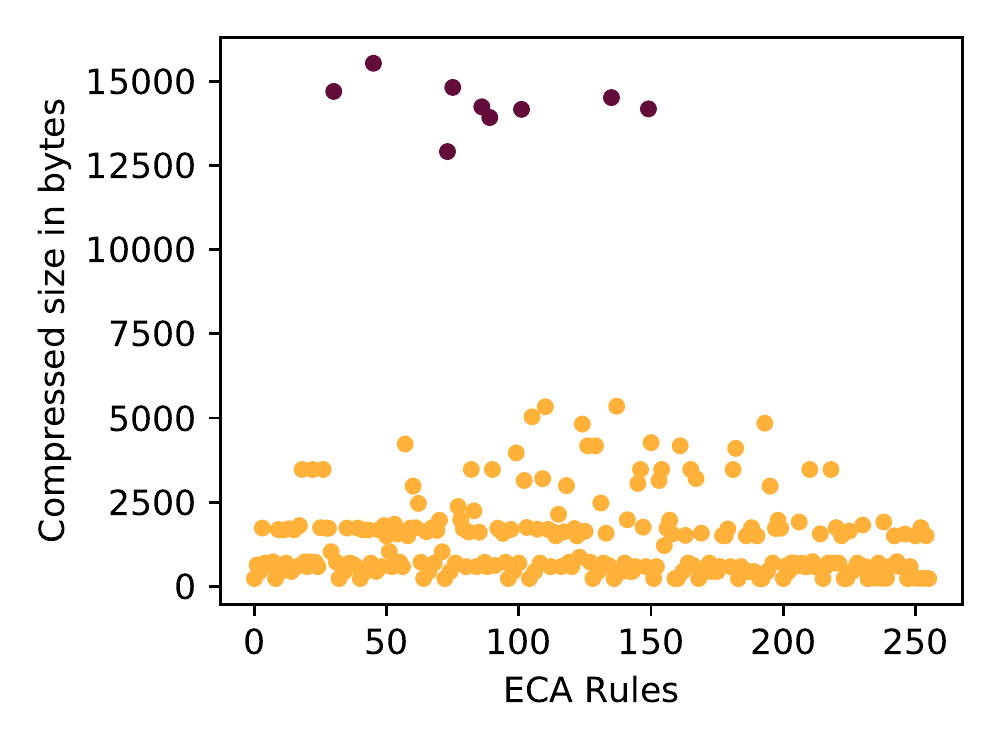}
  };
  \node at (3.9, 0) {
     \includegraphics[width=0.45\linewidth]{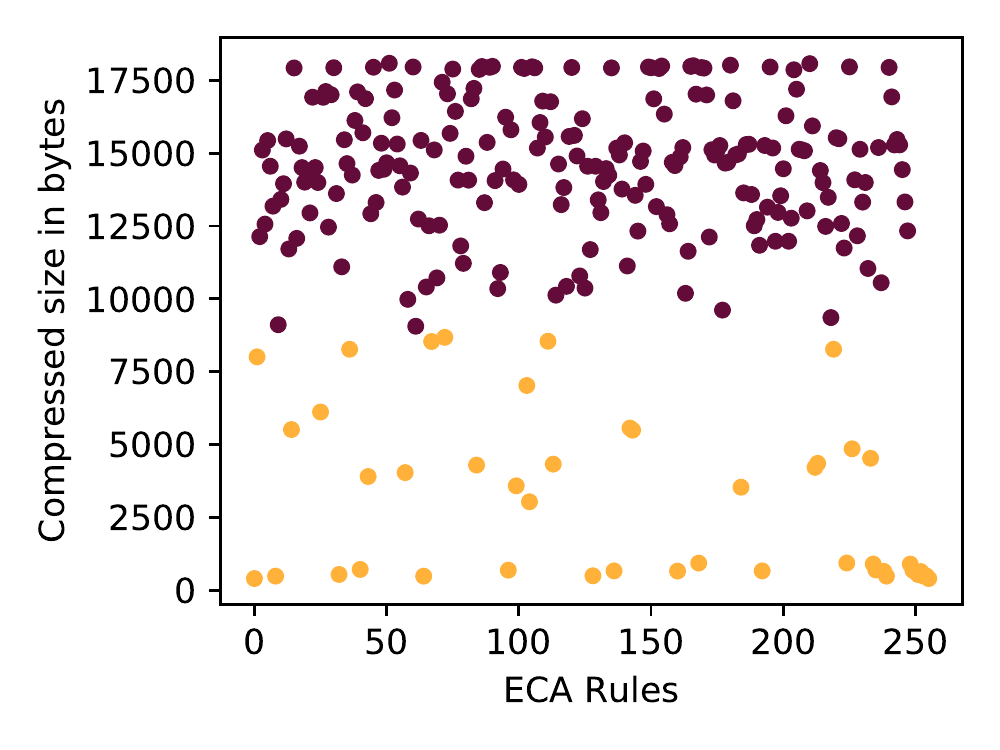}
 };
\end{tikzpicture}
\caption{Graphs representing the results of Zenil's method when different parameter values were used. They demonstrate the possible sensitivity of the results are. On the left, the ECAs were simulated for longer time, which caused complex rules 110, 124, 137, and 193 to no longer belong to the ``interesting'' purple cluster. On the right, the ECAs were simulated from a fixed, randomly chosen initial condition. In such case, we obtain entirely different clusters.}
\label{zenil_counter}
\end{figure}

His method nicely formalizes Wolfram's observations of the space-time diagrams. However, the results depend on the choice of initial conditions as well as the grid size, data representation, and the compression algorithm. We conducted multiple experiments presented in Figure \ref{zenil_counter}, which suggest that Zenil's results might be sensitive to the choice of such parameters. We note that he addresses the sensitivity to the choice of the initial configurations in the second part of his paper (\cite{zenil_compression}). 

In vast CA spaces where it is not feasible to examine every CA and mark it into one of Wolfram's classes by hand, it would not be clear how the parameter values should be chosen. Moreover, the data representation causes the extension of this method to more general dynamical systems to be problematic; for example, using gzip to compress space-time diagrams of a 2D cellular automaton is suboptimal.

\hyphenation{ap-proximated}

\subsubsection{Wuensche's Z-parameter}
\cite{wuensche_global} chose an interesting approach by studying the ECA's behavior when reversing the simulations and computing the preimages of each configuration. He introduces the Z-parameter, representing the probability that a partial preimage can be uniquely prolonged by one symbol, and suggests that Class 4 CAs typically occurs at Z $\approx 0.75$. However, no clear classification is formed. The crucial advantage is that the Z-parameter depends only on the CA's local rule and can be computed effectively. It is, however, questionable whether studying only the local rule could describe the overall dynamics of a system sufficiently well.

We note that transients of CAs have been examined, as in \cite{wuensche_global} or \cite{gutowitz}. However, we are not aware of an attempt to compare the asymptotic growth of transients for different ECA.

\subsection{Transient Classification of ECA}
For each ECA given by a local rule $f$, we consider the sequence of systems
$$D_3= (\{0, 1 \}^3, F_3), D_4= (\{0, 1 \}^4, F_4), \ldots$$
which represent the ECAs operating on grids of growing size. We can apply the transient classification to this sequence, as described in Section \nameref{general_method} to estimate the asymptotic growth of the average transient lengths for each ECA.

We consider all 256 ECAs up to equivalence classes obtained by changing the role of ``left'' and ``right'' neighbor, the role of 0 and 1 state, or both. 
It can be easily shown that automata in the same equivalence class have isomorphic phase spaces for any grid size. Thus, they perform the same computation. This yields 88 effectively different ECAs, each being a representative with the minimum Wolfram number from its corresponding equivalence class.
In this section, we present the classification of the 88 unique ECAs based on their asymptotic transient growth.

\hyphenation{ap-proximated}

\subsubsection{Results}

We obtained a surprisingly clear classification of all the 88 unique ECAs with four major classes corresponding to the bounded, logarithmic, linear, and exponential growth of average transients. Below, we give a more detailed description of each class.

\begin{figure}[h!] 
\begin{tikzpicture}[thick, every node/.style={inner sep=0,outer sep=0}]
  \node at (4, +0.08) {
     \includegraphics[width=0.44\linewidth]{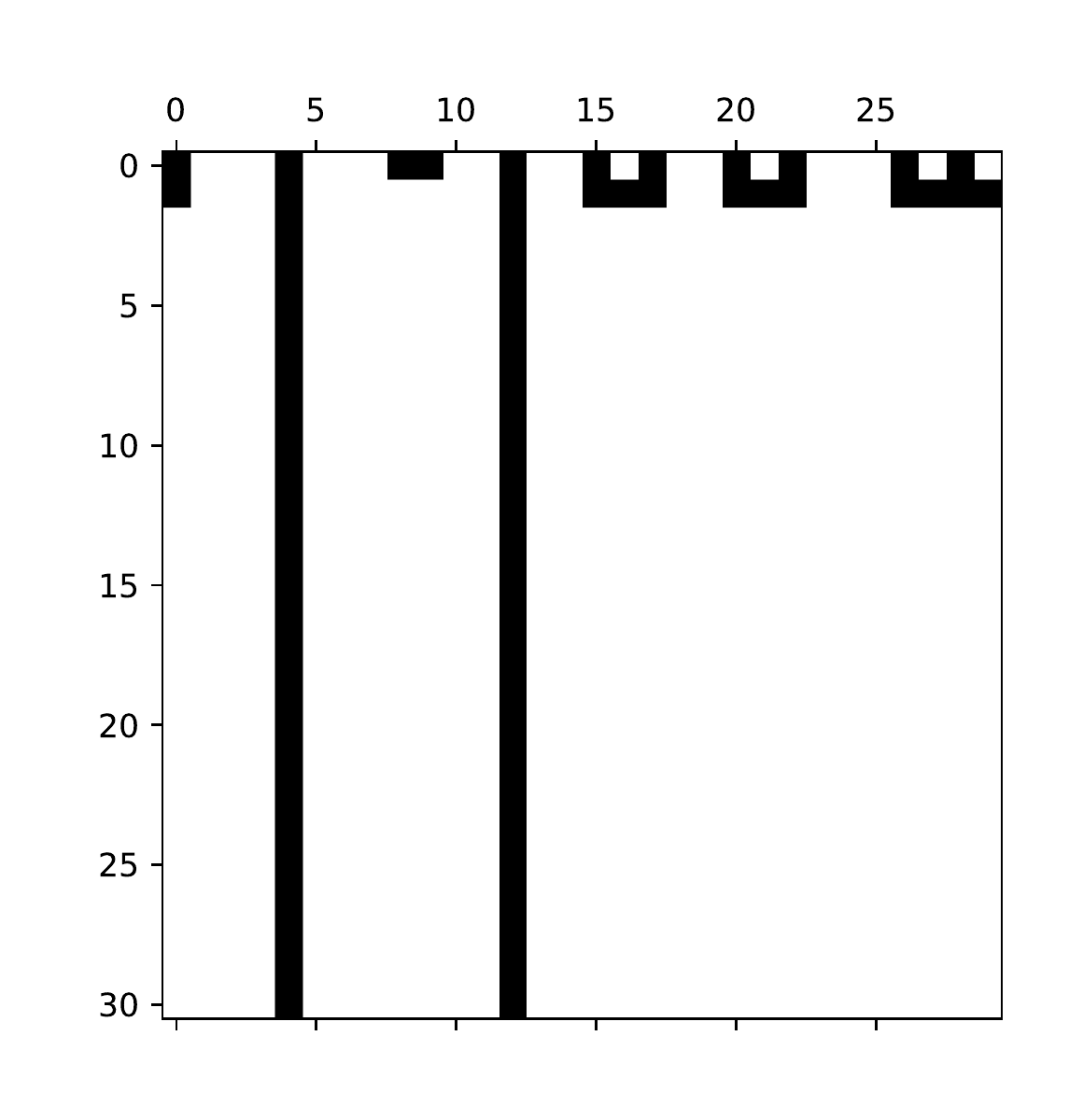}
  };
  \node at (-0.1, -0.1) {
     \includegraphics[width=0.6\linewidth]{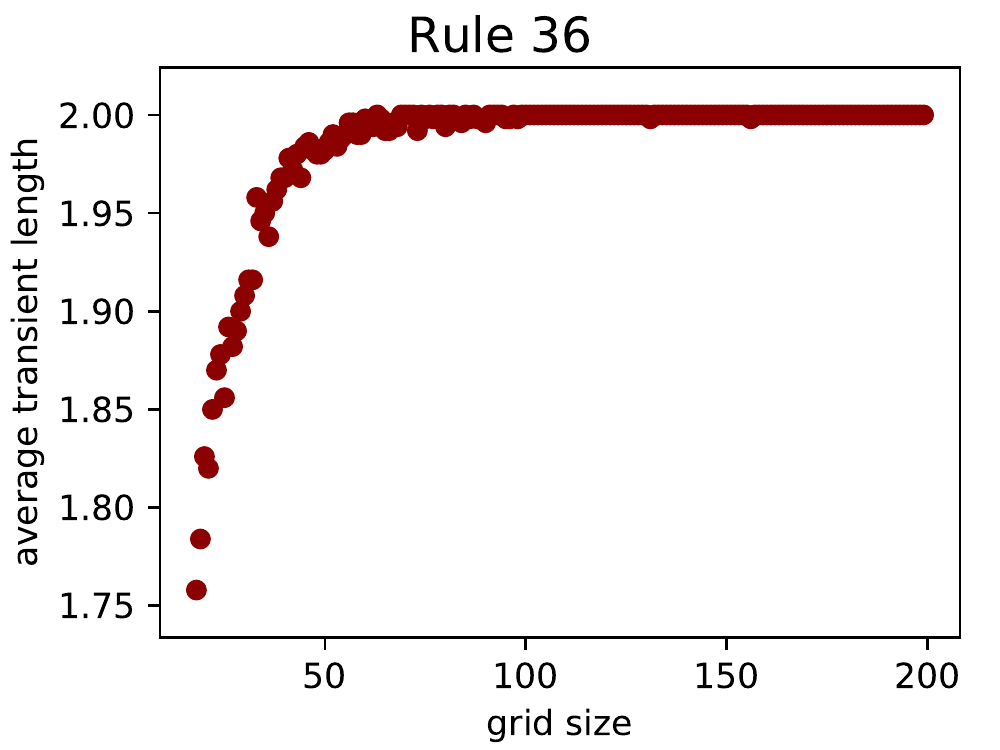}
  };
\end{tikzpicture}
\caption{Bounded Class rule 36. The average transient plot is on the left, the space-time diagram on the right.}
\label{bounded}
\end{figure}

\paragraph*{Bounded Class:} 27/88 rules ($30.68\%$). The average transient lengths were bounded by a constant independent of the grid size. This suggests that the long term dynamics of such automata can be predicted efficiently. See Figure \ref{bounded}.

\begin{figure}[h!] 
    \centering
    \begin{tikzpicture}[thick, every node/.style={inner sep=0,outer sep=0}]
  \node at (4, +0.08) {
     \includegraphics[width=0.44\linewidth]{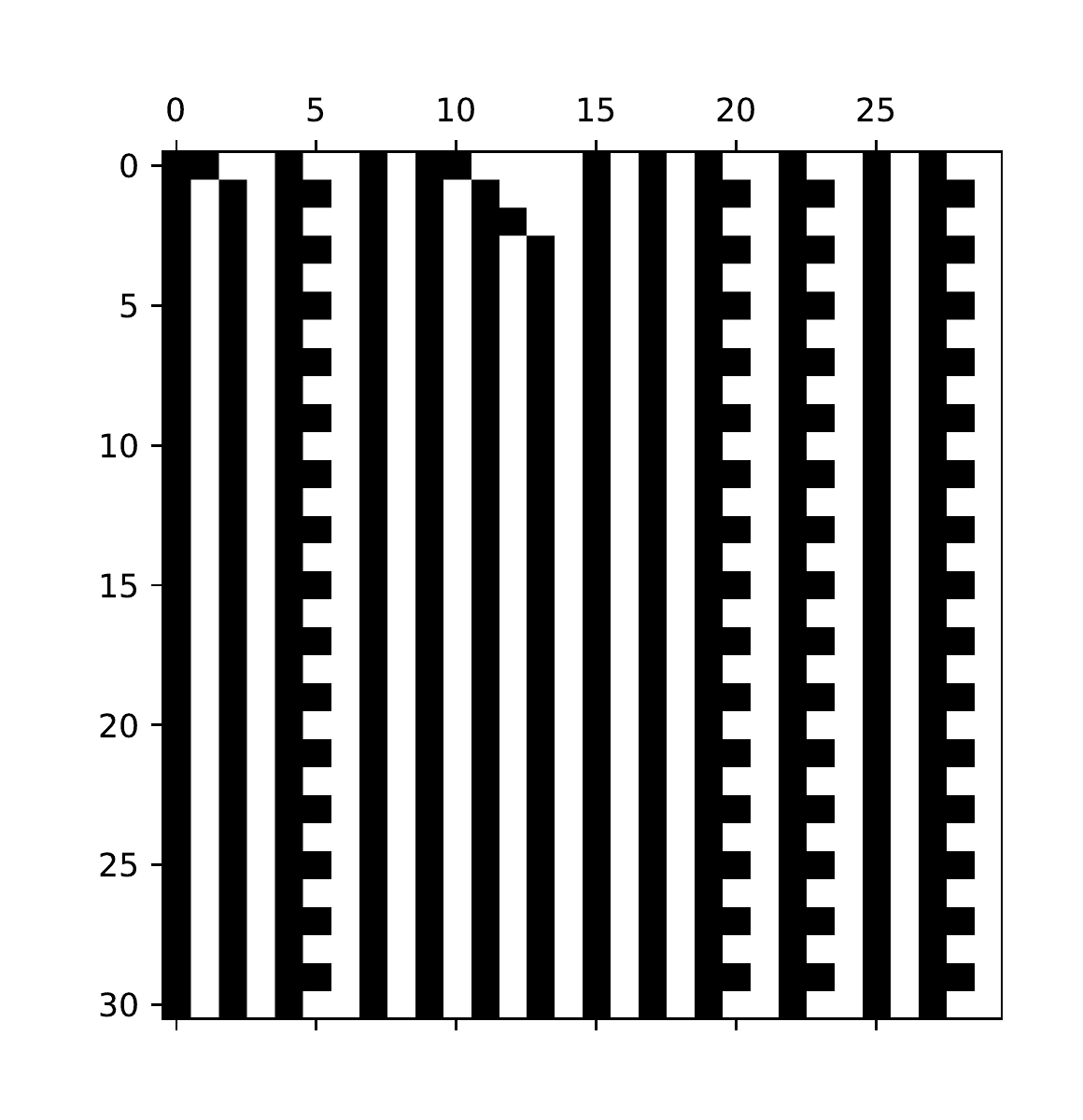}
  };
  \node at (-0.1, -0.1) {
     \includegraphics[width=0.6\linewidth]{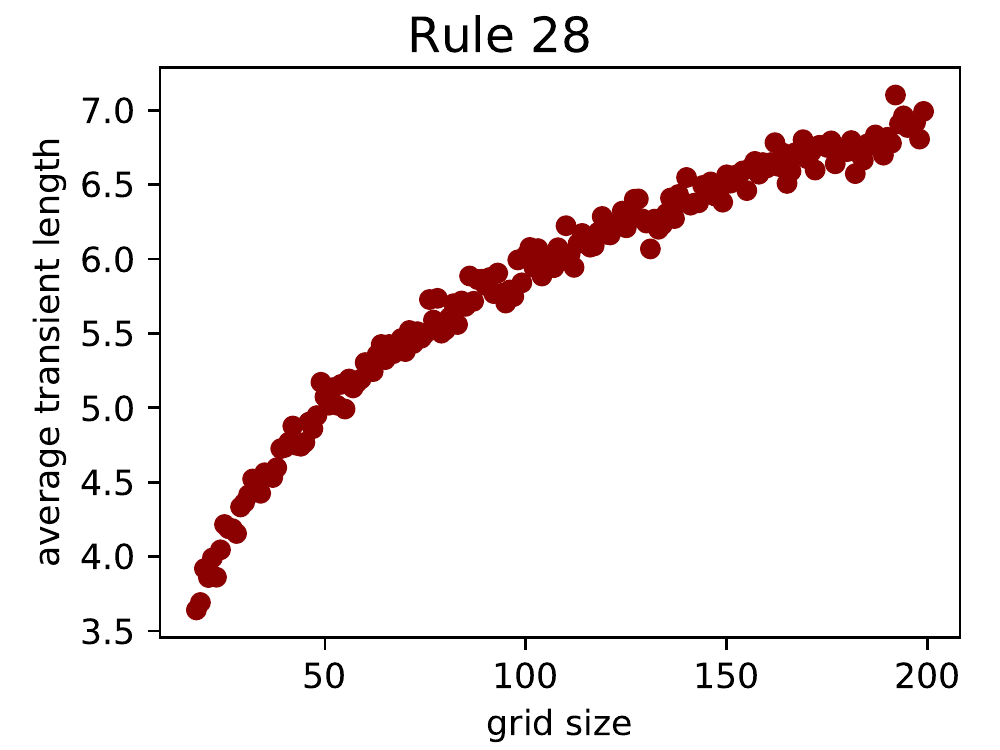}
  };
\end{tikzpicture}
\caption{Log Class rule 28. The average transient plot is on the left, the space-time diagram on the right.}
\label{log}
\end{figure}

\paragraph*{Log Class:}39/88 rules ($44.32\%$). The largest ECA class exhibits logarithmic average transient growth. The event of two cells at a large distance ``communicating'' is improbable for this class. See Figure \ref{log}.

{}

\paragraph*{Lin Class:}8/88 rules ($9.09\%$). On average, information can be aggregated from cells at an arbitrary distance. This class contains automata whose space-time diagrams resemble some sort of computation. This is supported by the fact that this class contains two rules known to have a nontrivial computational capacity: rule 184, which computes the majority of black and white cells, and rule 110, which is the only ECA so far proven to be Turing complete (\cite{cook}).

\begin{figure}[h!] 
\begin{tikzpicture}[thick, every node/.style={inner sep=0,outer sep=0}]
  \node at (4, +0.08) {
     \includegraphics[width=0.44\linewidth]{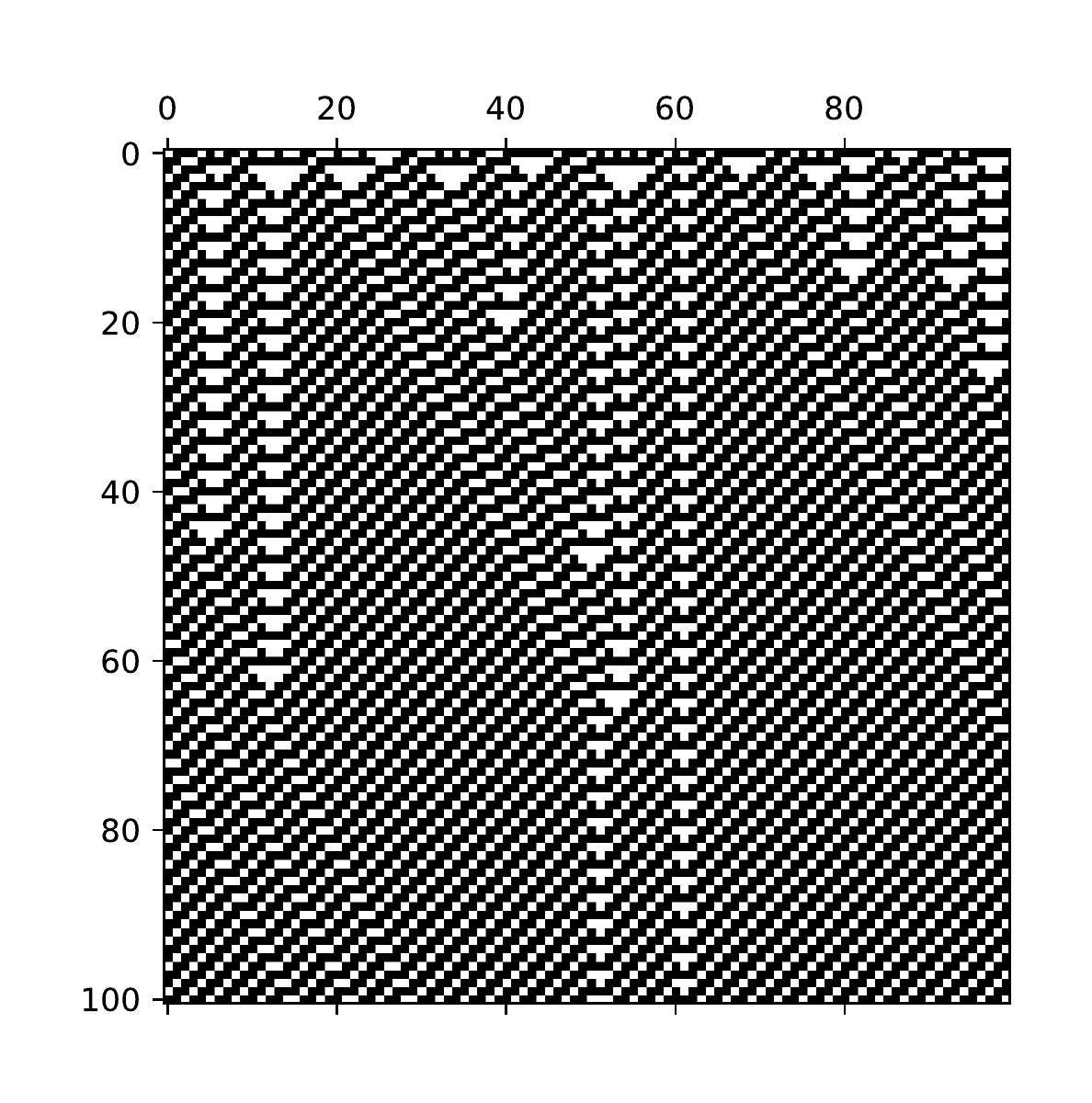}
  };
  \node at (-0.215, -0.1) {
     \includegraphics[width=0.6\linewidth]{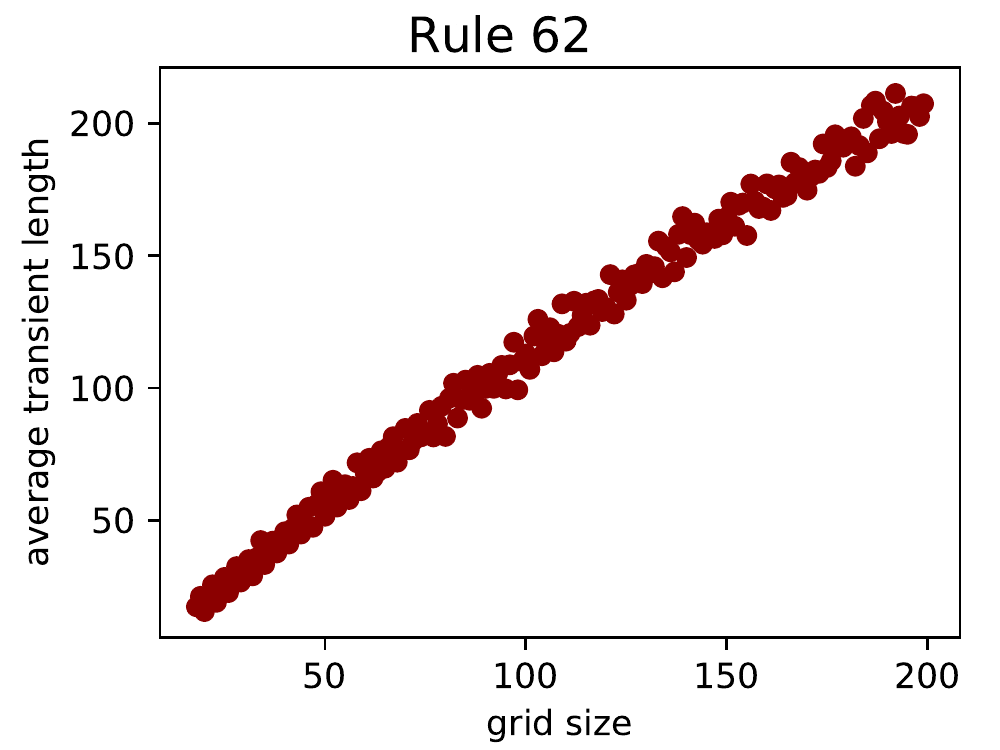}
  };
\end{tikzpicture}
\caption{Lin Class rule 62. The average transient plot is on the left, the space-time diagram on the right.}
\label{lin}
\end{figure}

We note that rules in this class are not necessarily complex as the interesting behavior seems to correlate with the slope of the linear growth. Most of the Class Lin rules had only a very gradual incline. In fact, the only two rules with such slope greater than 1, rules 110 and 62, seem to be the ones with the most interesting space-time diagrams. See Figure \ref{lin}.

We are aware that average transients of rules in Lin Class might turn out to grow logarithmically or exponentially given enough data samples. In such a case, the rules in Lin Class show a significantly slower convergence to their asymptotic behavior, which supports the hypothesis that they belong to a phase transition region.

\begin{figure}[h!] 
\begin{tikzpicture}[thick, every node/.style={inner sep=0,outer sep=0}]
  \node at (4, +0.08) {
     \includegraphics[width=0.44\linewidth]{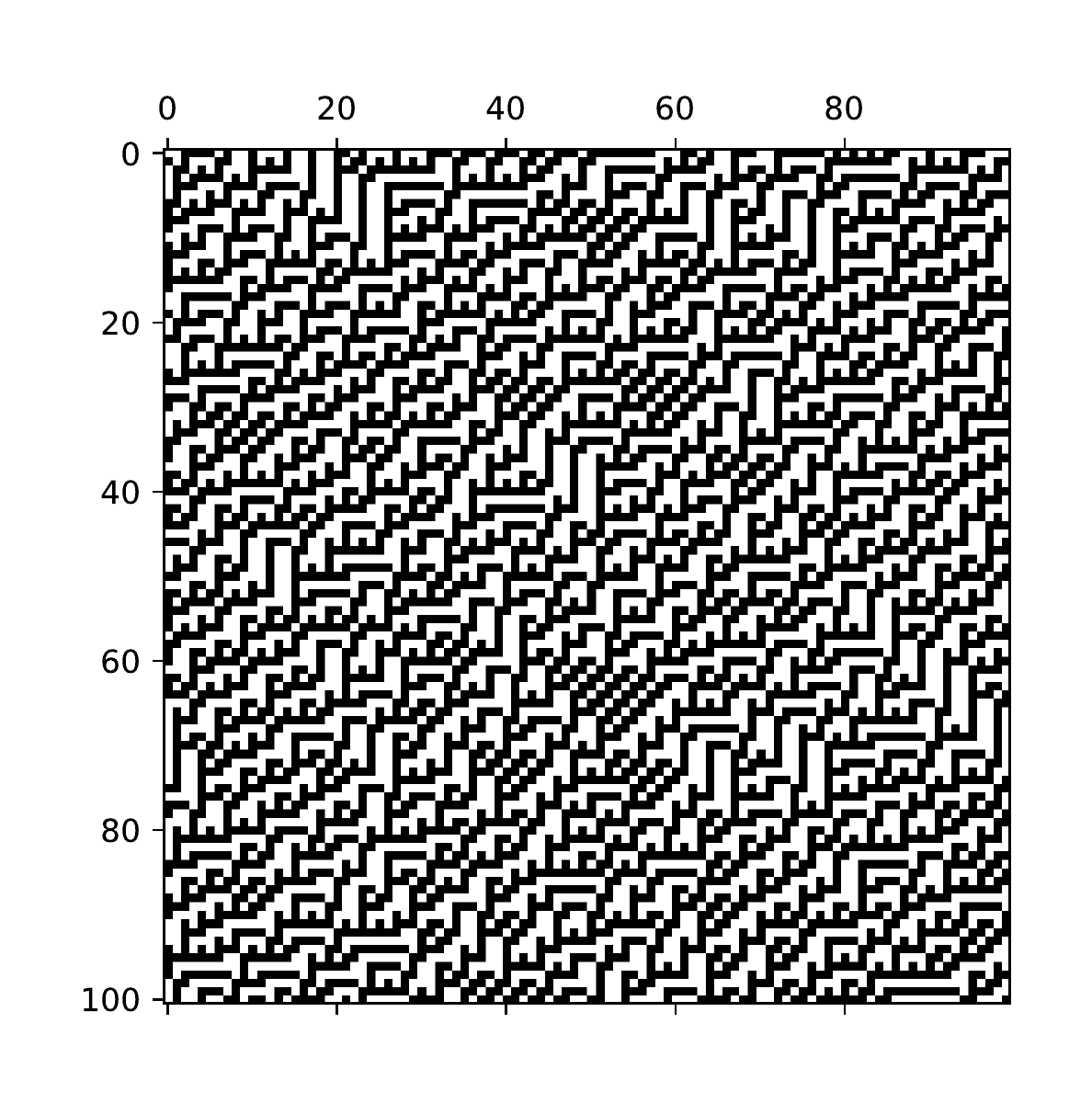}
  };
  \node at (-0.215, -0.1) {
     \includegraphics[width=0.6\linewidth]{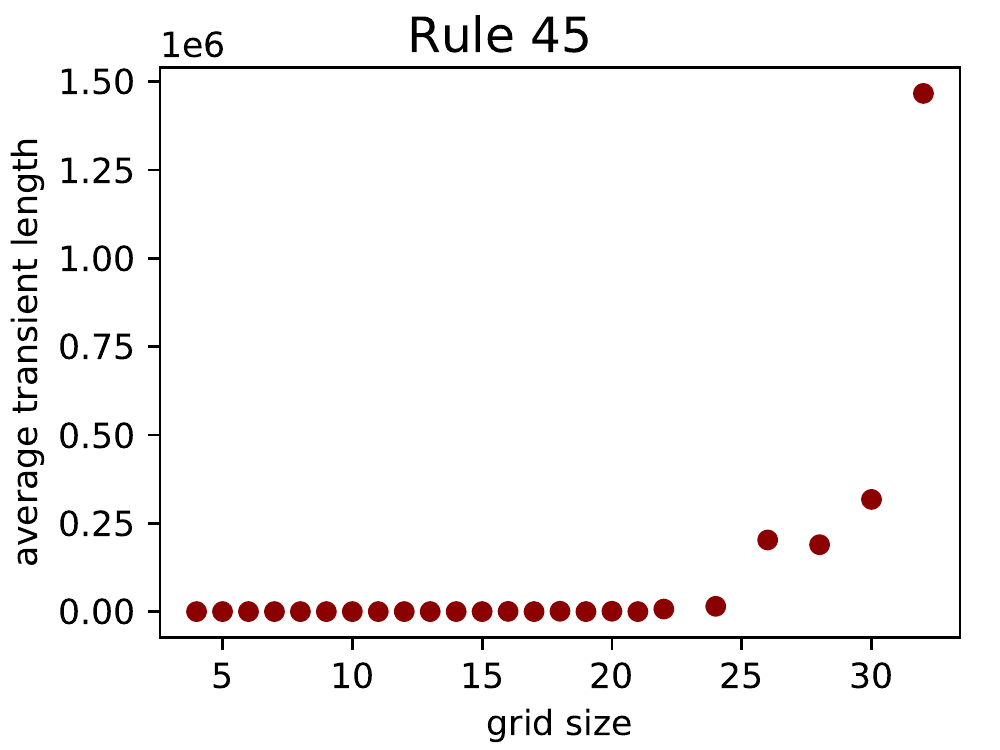}
  };
\end{tikzpicture}
\caption{Exp Class rule 45. The average transient plot is on the left, the space-time diagram on the right.}
\label{exp}
\end{figure}

\paragraph*{Exp Class:}6/88 rules ($6.82\%$). This class has a striking correspondence to automata with chaotic behavior. Visually, there seem to be no persistent patterns in the configurations. Not only the transients but also the attractor lengths are significantly larger than for other rules. The rules with the fastest growing transients are 45, 30 and 106. See Figure \ref{exp}.

\begin{figure}[h!] 
\begin{tikzpicture}[thick, every node/.style={inner sep=0,outer sep=0}]
  \node at (4, +0.08) {
     \includegraphics[width=0.44\linewidth]{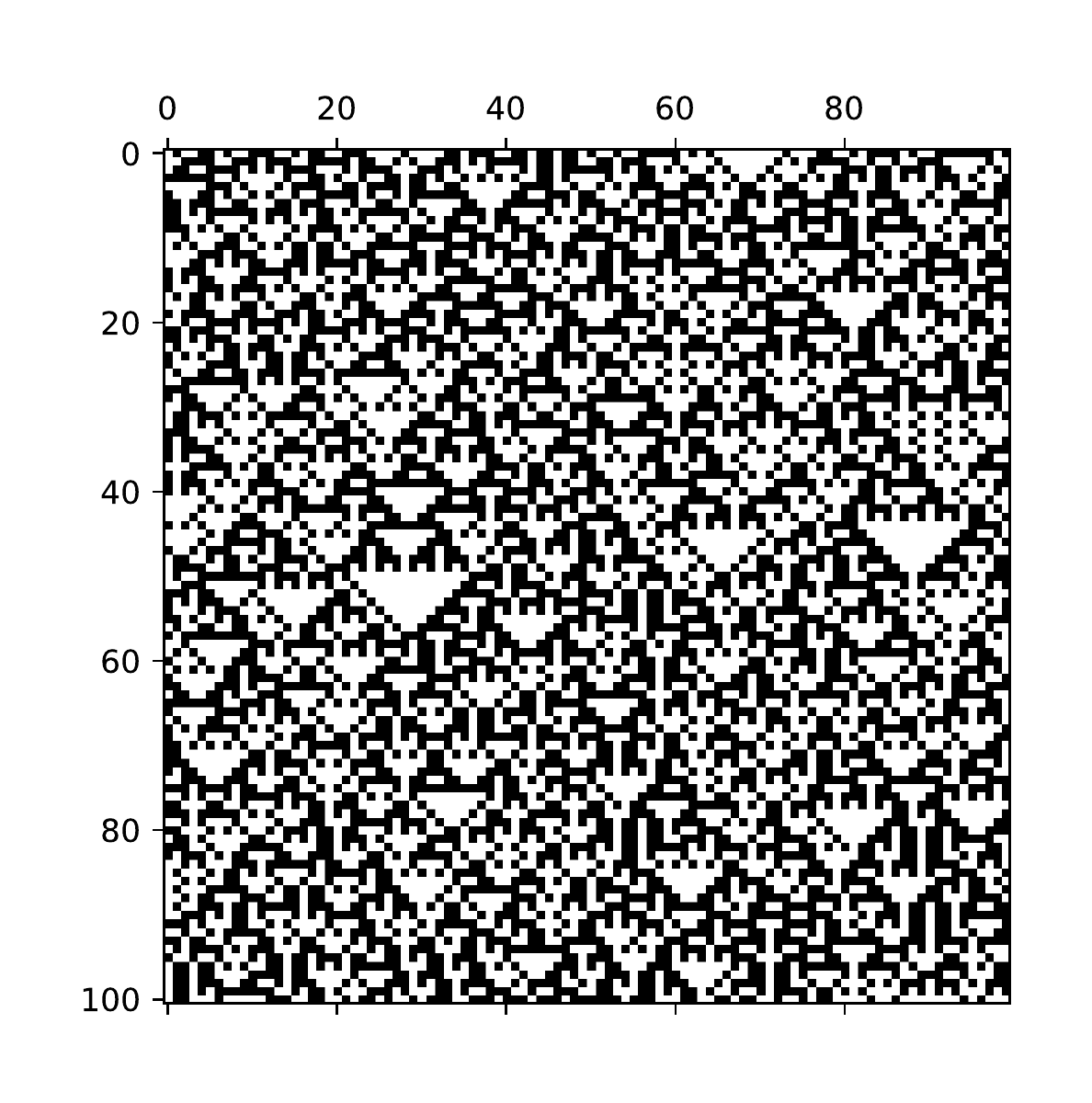}
  };
  \node at (-0.215, -0.1) {
     \includegraphics[width=0.6\linewidth]{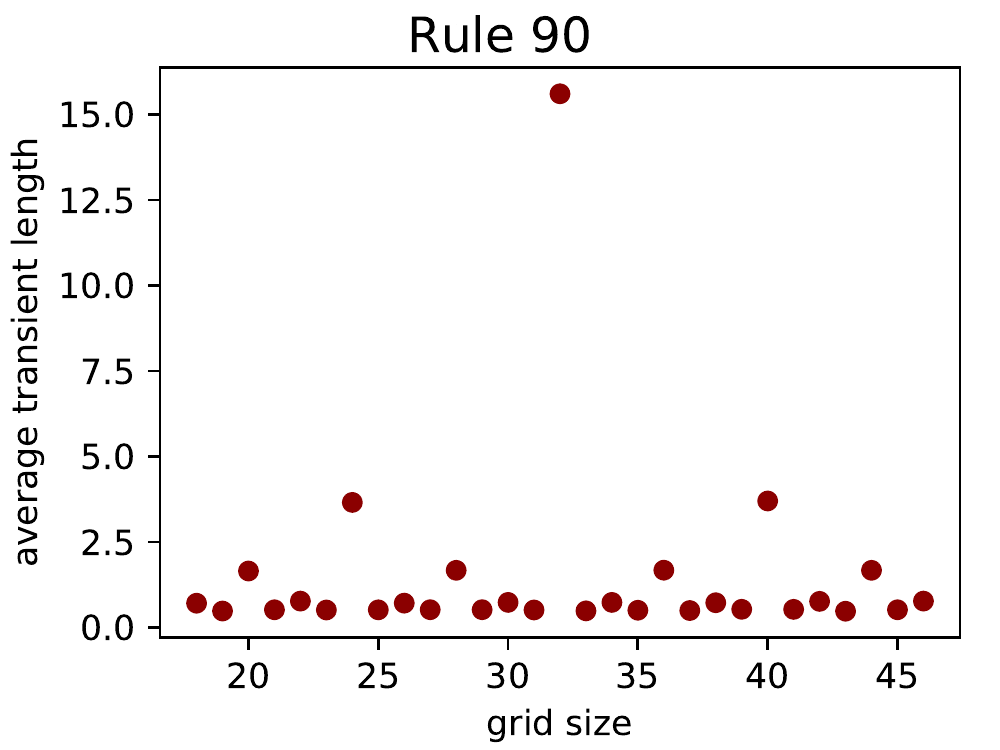}
  };
\end{tikzpicture}
\caption{Affine Class rule 90. The average transient plot is on the left, the space-time diagram on the right.}
\label{affine}
\end{figure}

\paragraph*{Affine Class:}4/88 rules ($4.55\%$). This class contains rules 60, 90, 105, and 150 whose local rules are affine Boolean functions. Such automata can be studied algebraically and predicted efficiently. It was shown in \cite{algCA} that the transient lengths of rule 90 depend on the largest power of 2, which divides the grid size. Therefore, the measured data did not fit any of the functions above but formed a rather specific pattern. See Figure \ref{affine}.

\begin{figure}[h!]
\begin{tikzpicture}[thick, every node/.style={inner sep=0,outer sep=0}]
  \node at (4, +0.08) {
     \includegraphics[width=0.44\linewidth]{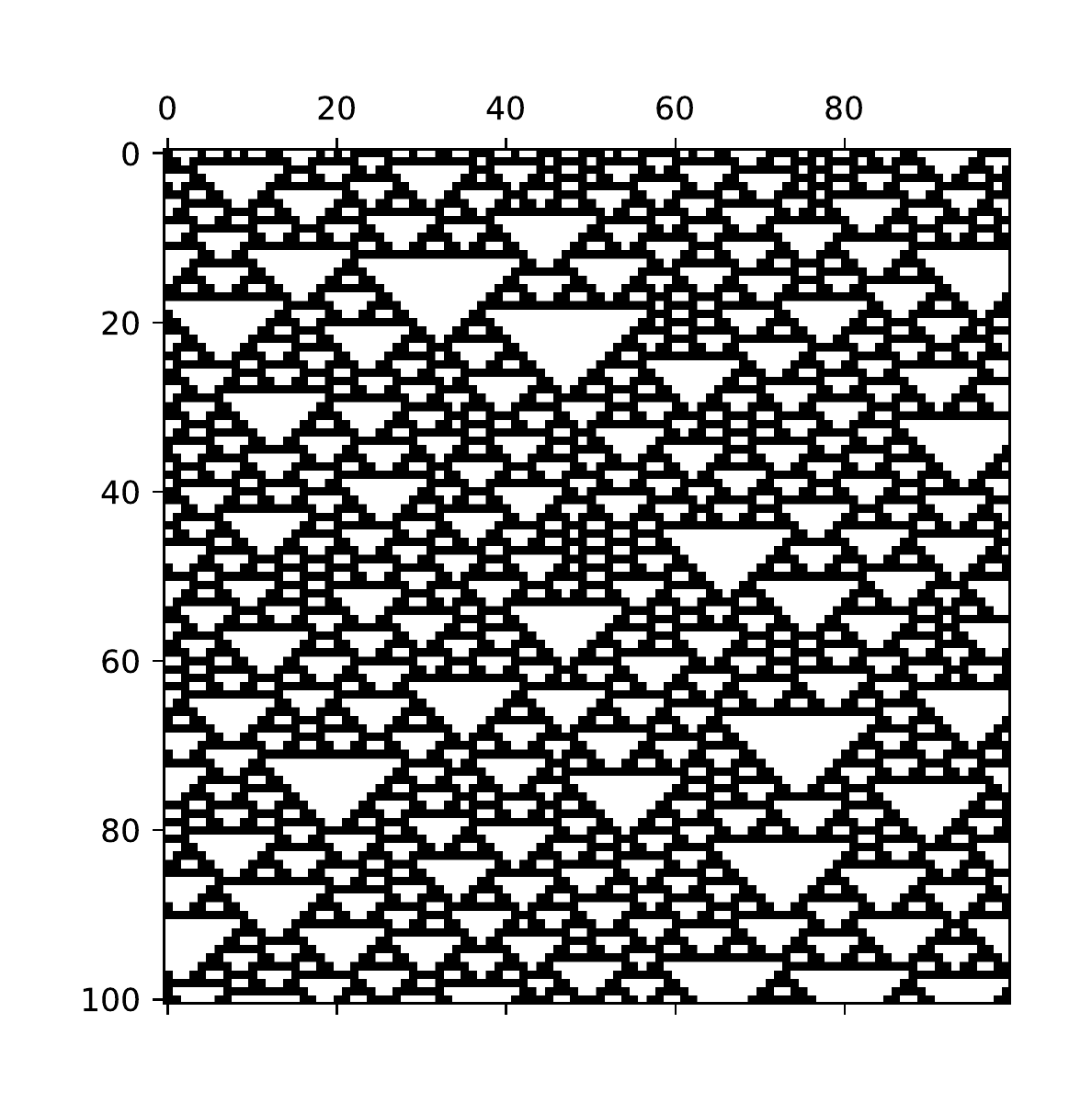}
  };
  \node at (-0.215, -0.1) {
     \includegraphics[width=0.6\linewidth]{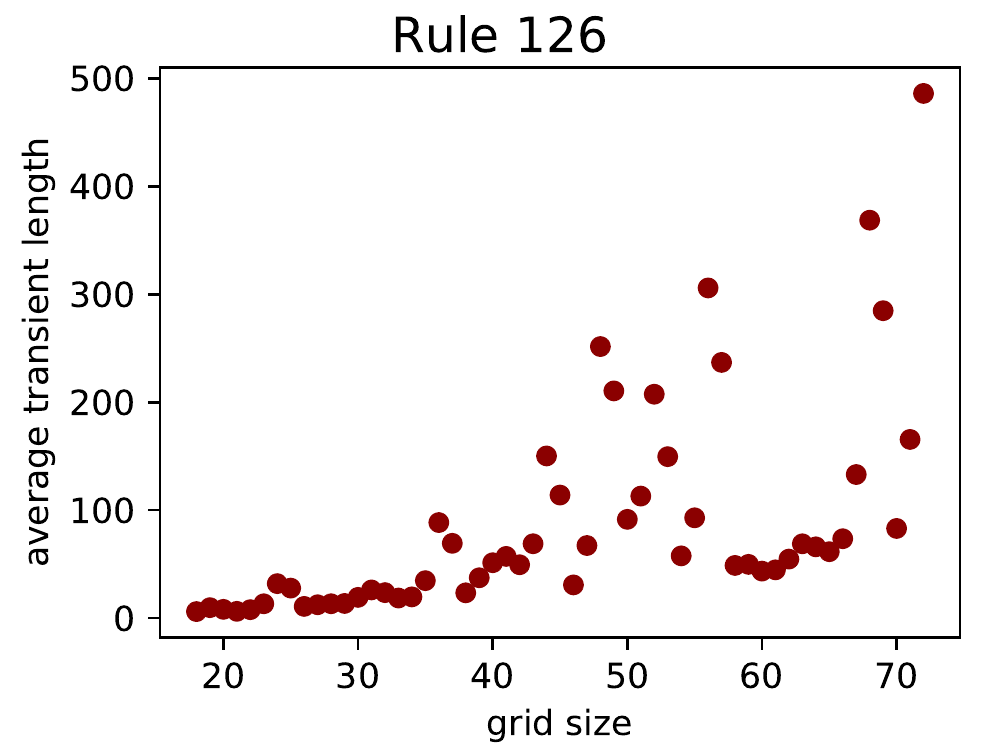}
  };

\end{tikzpicture}
\caption{Fractal Class rule 126. The average transient plot is on the left, the space-time diagram on the right.}
 \label{fractal}
\end{figure}

\paragraph*{Fractal Class:}4/88 rules  ($4.55\%$). This class contains rules 18, 122, 126, and 146 which are sensitive to initial conditions. Their evolution either produces a fractal structure resembling a Sierpinski triangle or a space-time diagram with no apparent structures. We could say such rules oscillate between easily predictable behavior and chaotic behavior. Their average transients and periods grow quite fast. See Figure \ref{fractal}.

\subsection{Discussion}
We have also tried to measure the asymptotic growth of the average attractor size $a_u$, $u \in \{0, 1 \}^n$ as well as the average rho value defined as $\rho_u = t_u + a_u$. This, however, produced data points, which could not be fitted to simple functions well. This is due to the fact that many automata have attractors consisting of a configuration, which is shifted by one bit to the left, resp. right, at every time step. The size of such an attractor then depends on the greatest common divisor of the size of the period of the attractor and the grid size, and this causes oscillations. We conclude that such phase-space properties are not suitable for this classification method.

 \tablefirsthead{\hline \multicolumn{5}{|c|}{\textbf{Classification Comparison}} \\ \hline
 \multicolumn{1}{|c||}{\textbf{ECA}} &  \multicolumn{1}{|c|}{\textbf{Transient}} &
  \multicolumn{1}{|c|}{\textbf{Wolfram}} & \multicolumn{1}{|c|}{\textbf{Zenil}} &  \multicolumn{1}{|c|}{\textbf{Wuensche}}
  \\ \hline}
   \tablehead{\hline \multicolumn{5}{|c|}{\textbf{Classification Comparison}} \\ \hline
 \multicolumn{1}{|c||}{\textbf{ECA}} &  \multicolumn{1}{|c|}{\textbf{Transient}} &
  \multicolumn{1}{|c|}{\textbf{Wolfram}} & \multicolumn{1}{|c|}{\textbf{Zenil}} &  \multicolumn{1}{|c|}{\textbf{Wuensche}}
  \\ \hline}
   \tabletail{\hline \multicolumn{5}{|r|}{{Continued in the subsequent column}} \\ \hline}
  \tablelasttail{\hline}
  \bottomcaption{Comparing classifications of the 88 unique ECA.}
  \label{tab1}
\begin{center}
\begin{strictsupertabular}[!h]{@{\extracolsep{\fill}} |m{.8cm}||m{1.2cm}|m{.8cm}|m{1cm}|m{1cm}|  }
0 & bounded & 1 &1 or 2  & 0\\
1 & bounded & 2 &1 or 2 & 0.25\\
2 & bounded & 2 & 1 or 2& 0.25\\
3 & bounded & 2 &1 or 2& 0.25\\
4 & bounded &2 &1 or 2& 0.25\\
5 & bounded & 2 & 1 or 2 & 0.5\\
6 & log & 2 & 1 or 2 & 0.5\\
7 & log & 2 &  1 or 2& 0.75\\
8 & bounded & 1&1 or 2& 0.25\\
9 & lin & 2 & 1 or 2 & 0.5\\
10 & bounded & 2 & 1 or 2 & 0.5\\
11 & log & 2 & 1 or 2 & 0.75\\
12 & bounded & 2 & 1 or 2 & 0.5\\
13 & log & 2 & 1 or 2 & 0.75\\
14 & lin & 2 &  1 or 2& 0.75\\
15 & bounded & 2 & 1 or 2 & 1\\
18 & fractal & 2/3 & 1 or 2 & 0.5\\
19 & bounded & 2 &  1 or 2& 0.625\\
22 & exp & 2/3 & 1 or 2 & 0.75\\
23 & log & 2 & 1 or 2 & 0.5\\
24 & bounded & 2 &  1 or 2& 0.5\\
25 & lin & 2 & 1 or 2 & 0.75\\
26 & log & 2 &  1 or 2& 0.75\\
27 & log & 2 & 1 or 2 & 0.75\\
28 & log & 2 &  1 or 2& 0.75\\
29 & bounded & 2 & 1 or 2 & 0.5\\
30 & exp & 3 &  3 & 1\\
32 & log & 1 & 1 or 2 & 0.25 \\
33 & log & 2 &  1 or 2& 0.5\\
34 & bounded & 2 & 1 or 2 & 0.5\\
35 & log & 2 &  1 or 2& 0.625\\
36 & bounded & 2 & 1 or 2 & 0.5\\
37 & log & 2 & 1 or 2 & 0.75\\
38 & bounded & 2 & 1 or 2 & 0.75\\
40 & log & 1 &  1 or 2& 0.5\\
41 & log & 2 & 1 or 2 & 0.75\\
42 & bounded & 2 & 1 or 2 & 0.75\\
43 & lin & 2 & 1 or 2 & 0.5\\
44 & log & 2 & 1 or 2 & 0.75\\
45 & exp & 3 &  3 & 1\\
46 & bounded & 2 & 1 or 2 & 0.5\\
50 & log & 2 & 1 or 2 & 0.625\\
51 & bounded & 2 &1 or 2 & 1\\
54 & exp & 2/4 &  1 or 2 & 0.75\\
56 & log & 2 & 1 or 2 & 0.75\\
57 & lin & 2 & 1 or 2 & 0.75\\
58 & log & 2 & 1 or 2 & 0.75\\
60 & affine & 2 & 1 or 2 & 1\\
62 & lin & 2 & 1 or 2 & 0.75\\
72 & bounded & 1 & 1 or 2 & 0.5\\
73 & exp & 3/4 & 3 & 0.75\\
74 & log & 2 & 1 or 2 & 0.75\\
76 & bounded & 2 &1 or 2 & 0.625\\
77 & log & 2 & 1 or 2& 0.5\\
78 & log & 2 & 1 or 2& 0.75\\
90 & affine & 2 &1 or 2 & 1\\
94 & log & 2 & 1 or 2 & 0.75\\
104 & log & 1 & 1 or 2& 0.75\\
105 & affine & 2 &1 or 2 & 1\\
106 & exp & 3  &1 or 2 & 1\\
108 & bounded &1&1 or 2 & 0.75\\
110 & lin & 4 & 4 & 0.75\\
122 & fractal & 2/3 & 1 or 2& 0.75\\
126 & fractal & 2/3 & 1 or 2& 0.5\\
128 & log & 1 & 1 or 2 & 0.25\\
130 & log & 2 & 1 or 2& 0.5\\
132 & log & 2&1 or 2& 0.5\\
134 & log & 2 &1 or 2& 0.75\\
136 & log & 1 &1 or 2& 0.5\\
138 & bounded & 2 &1 or 2 & 0.75\\
140 & log & 2 &1 or 2 & 0.625\\
142 & lin & 2 &1 or 2 & 0.5\\
146 & fractal & 2/3 &1 or 2 & 0.75\\
150 & affine & 2 &1 or 2 & 1\\
152 & log & 2 & 1 or 2 & 0.75\\
154 & bounded & 2/3  & 1 or 2& 1\\
156 & log & 2 &1 or 2 & 0.75\\
160 & log & 1 & 1 or 2 & 0.5\\
162 & log & 2 & 1 or 2& 0.75\\
164 & log & 2 & 1 or 2& 0.75\\
168 & log & 1 & 1 or 2 & 0.75\\
170 & bounded & 2 & 1 or 2& 1\\
172 & log & 2 & 1 or 2 & 0.75\\
178 & log & 2 & 1 or 2& 0.5\\
184 & lin & 2 & 1 or 2& 0.5\\
200 & bounded & 1 & 1 or 2 & 0.625\\
204 & bounded & 2 &1 or 2 & 1\\
232 & log & 1 & 1 or 2& 0.5\\
  \end{strictsupertabular}
\end{center}

Exhaustive comparison for each ECA is presented in Table \ref{tab1}. 

\paragraph{Wolfram's Classification -- Discussion}
The significance of our results for ECAs stems precisely from the fact that the transient classification corresponds to Wolfram's so well. As it is not clear for many rules which Wolfram class they belong to, the main advantage is that we provide a formal criterion upon which this could be decided.

In particular, rules in Classes Bounded and Log correspond to rules in either Class 1 or 2. Class Exp corresponds to the chaotic Class 3, and Class Lin contains Class 4 together with some Class 2 rules. We mention an interesting discrepancy: rule 54, which is possibly considered by Wolfram to be Turing complete, belongs to the Class Exp. This might suggest that computations performed by this rule can be on average quite inefficient.

\hyphenation{ap-proximated}
\paragraph{Zenil's Classification -- Discussion}
Zenil's Classification of ECAs offers a great formalization of Wolfram's and seems to roughly correspond to it. Compared to the transient classification, it is, however, less fine-grained. Moreover, it contains some arbitrary parameters, such as the data representation and compression algorithm used. In addition, it uses a clustering technique, which requires data of multiple automata to be mutually compared in order to give rise to different classes. In contrast, the transient class can be determined for a single automaton without any context.

Another important difference is that Zenil observed the simulations from a fixed initial configuration; i.e., he examined the local dynamics of ECA. In contrast, the transient classification is studying their global dynamics.

\hyphenation{ap-proximated}
\paragraph{Wuensche's Z-parameter -- Discussion}
Wuensche suggests that complex behavior occurs around $Z = 0.75$, which agrees with the fact that Lin Class rules with a steep slope (rule 110, 62, and 25) have this $Z$ value precisely. However, the $Z = 0.75$ is in fact quite frequent. This suggests that thanks to its simplicity, the $Z$ parameter can be used to narrow down a vast space of CA rules when searching for complexity. However, more refined methods have to be subsequently applied to find concrete CAs with interesting behavior.
\hyphenation{ap-proximated}

\subsection{Transient Classification of 2D CA}
So far, we have examined the toy model of ECA. Transient classification’s true usefulness would stem from its application to more complex CAs, where it could be used to discover automata with interesting behavior.

Therefore, we applied the classification on a subset of two-dimensional CAs with a $3 \times 3$ neighborhood and three states to see whether 2D automata would still exhibit such clear transient growths.

We work with 2D CAs operating on a finite square grid of size $n \times n$. We consider the topology of the grid to be that of a torus for each cell to have a uniform neighborhood. We estimated the average transient length and measured the asymptotic growth with respect to $n$ (i.e., the size of the square grid's side). This is motivated by the fact that in a $n \times n$ grid, the greatest distance between two cells depends linearly on $n$ rather than quadratically.

To reduce the vast automaton space, we only considered such automata whose local rules are invariant to all the symmetries of a square. As there are still $3^{2861}$ such symmetrical 2D CAs, we randomly sampled 10 000 of them.

For such a large space, we cannot examine each CA individually. Therefore, we fit the average transient growth to bounded, logarithmic, linear, polynomial, and exponential functions to obtain the classes Bounded, Log, Lin, Poly, and Exp. If none of the fits gives a good enough score (i.e., $R^2 > 85\%$), then we mark the corresponding CAs as unclassified.
We were able to classify $93.03 \%$ of 10 000 sampled automata with a time bound of 40 seconds for the computation of one transient length value on a single CPU. We estimate that most CAs are unclassified due to such computation resources restriction or rather strict conditions we imposed on a good regression fit. In this large space of 2D CAs, the Exp Class seems to dominate the rule space. Another interesting aspect in which 2D CAs differ from the ECAs is the emergence of rules in the Poly Class; the transients of such rules grow approximately quadratically. Moreover, our results suggest that the occurrence of Bounded Class CAs in 2D is much scarcer as we found no such CA in our sample. See Table \ref{2D}.

\begin{table}[h!]
\centering
\begin{tabular}{ |m{3cm}||m{4.6cm}|  }
 \hline
 \multicolumn{2}{|c|}{Classification of 2D 3-state CAs (10 000 samples)} \\
 \hline
 \centering
 Transient Class & \centering Percentage of CA
 \tabularnewline
 \hline
Bounded Class & 0\%\\
Log Class &  18.21\%\\
Lin Class & 1.17\%\\
Poly Class & 1.03\%\\
Exp Class & 72.62\%\\
Unclassified & 6.97\%\\
 \hline
\end{tabular}
\caption{   Classification of 10 000 randomly sampled symmetric 2D 3-state CA.}
\label{2D}
\end{table}

We observed the space-time diagrams of randomly sampled automata from each class to infer its typical behavior. On average, the Log Class automata quickly enter attractors of small size. Lin Class exhibits the emergence of various local structures. For automata with a more gradual incline, such structures seem to die out quite fast. Automata with steeper slopes exhibit complex interactions of such structures. The Poly class automata with a steep slope seem to produce spatially separated regions of chaotic behavior against a static background. In the case of more gradual slopes, some local structures emerge. Finally, the Exp Class seems to be evolving chaotically with no apparent local structures. We present various examples of CA evolution dynamics in the form of GIF animations here\footnotemark.
\footnotetext{\url{http://bit.ly/trans_class}}

This suggests that the region of Lin Class with a steep slope and Poly class with a more gradual incline seems to contain a non-trivial ratio of automata with complex behavior. In this sense, the transient classification can assist us to automatically search for complex automata similarly to the method designed by \cite{hugo} where interesting novel automata were discovered by measuring growth of structured complexity using a data compression approach.

\subsection{Transients Classification of Other Well Known CA}
We were interested whether some well-known complex automata from larger CA spaces would conform to the transient classification as well. As we show in this section, the result is positive.

\begin{figure}[h!] 
\begin{tikzpicture}[thick, every node/.style={inner sep=0,outer sep=0}]
  \node at (4, +0.08) {
     \includegraphics[width=0.44\linewidth]{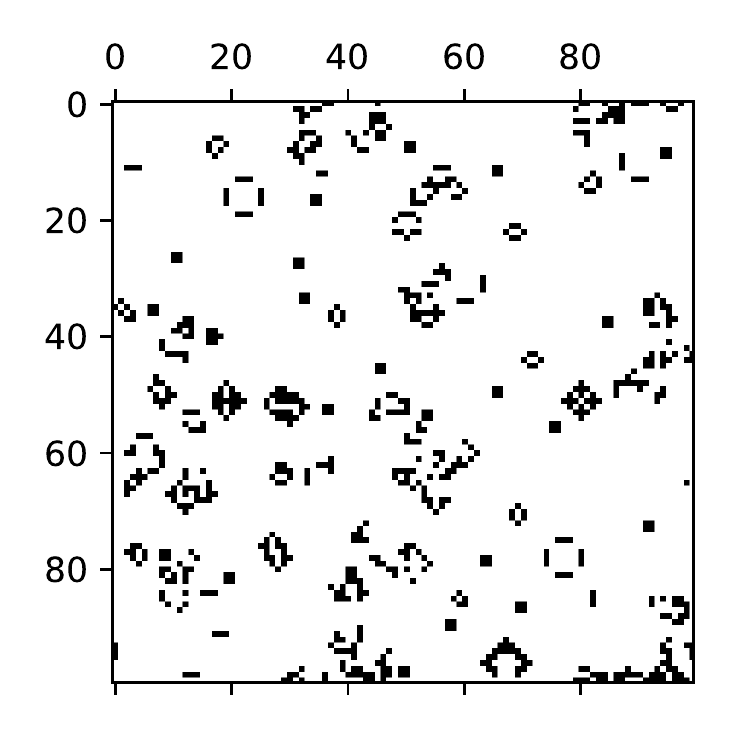}
  };
  \node at (-0.215, -0.1) {
     \includegraphics[width=0.6\linewidth]{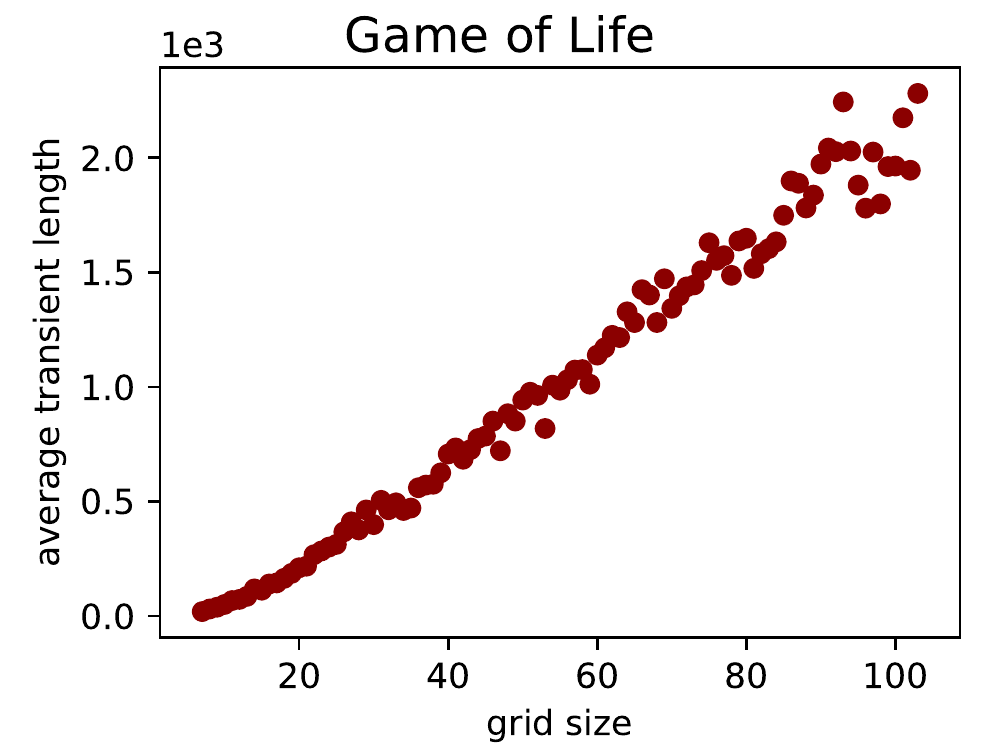}
  };
\end{tikzpicture}
\caption{Game of Life. The average transient growth plot is on the left. On the right, we show a space-time diagram at time $t=200$ started from a random initial configuration.}
\label{gol_pic}
\end{figure}

\paragraph{Game of Life} As the left plot in Figure \ref{gol_pic} suggests, the Turing complete Game of Life (\cite{game_of_life}) seems to fit the Lin Class. This is confirmed by the linear regression fit with $R^2 \approx 98.4\% $.

\begin{figure}[h!]
\begin{tikzpicture}[thick, every node/.style={inner sep=0,outer sep=0}]
  \node at (4, +0.08) {
     \includegraphics[width=0.44\linewidth]{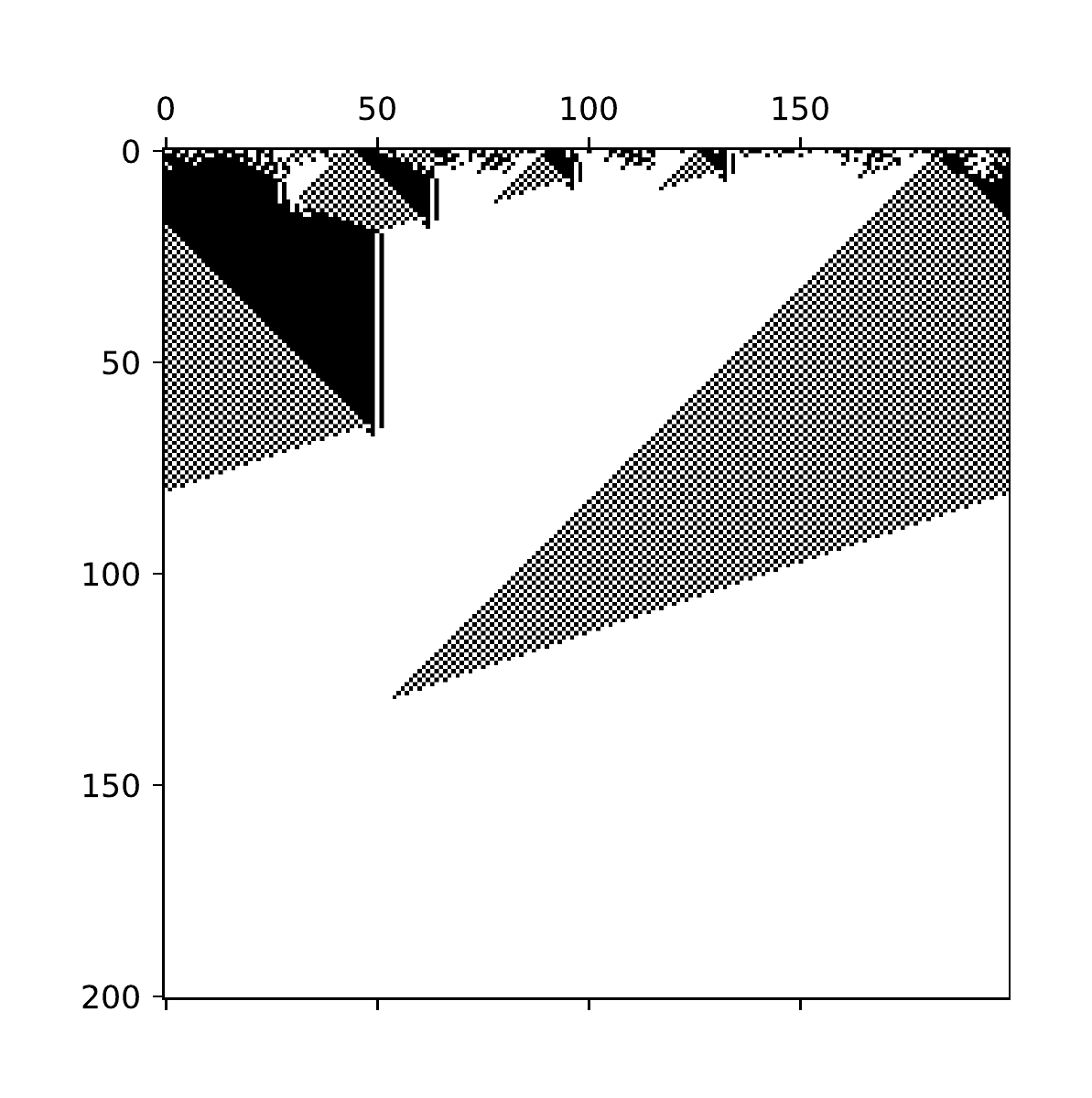}
  };
  \node at (-0.215, -0.1) {
     \includegraphics[width=0.6\linewidth]{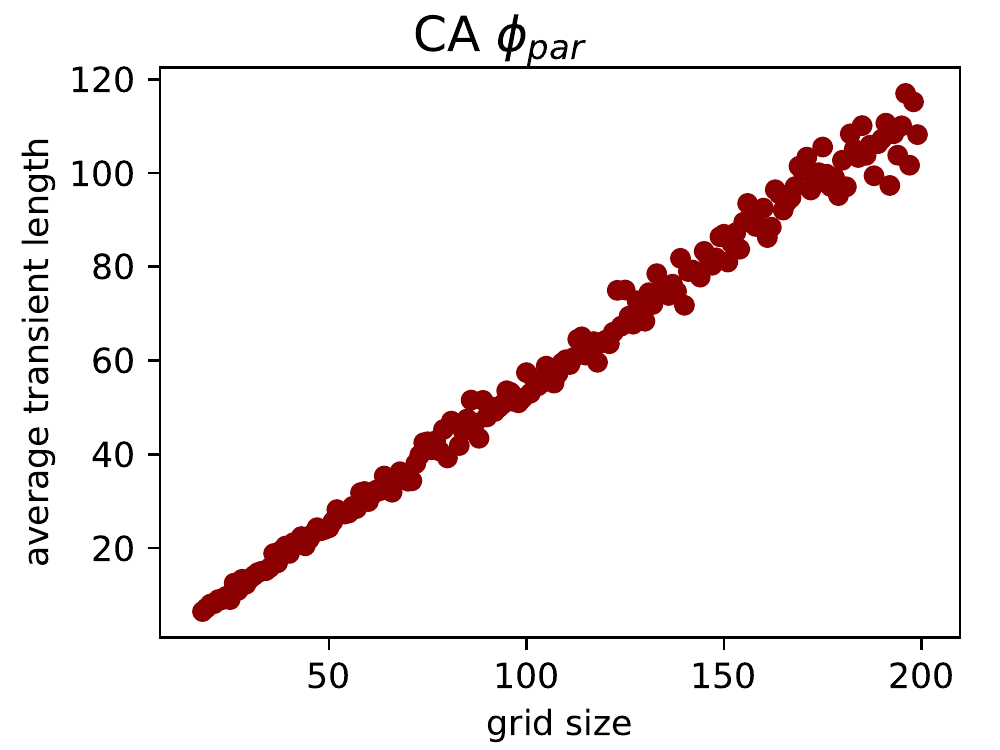}
  };

\end{tikzpicture}
\caption{Cellular automaton $\phi_{par}$. The average transient growth plot is on the left. On the right, we show a space-time diagram simulated from a random initial configuration.}
\label{phi_par}
\end{figure}

\paragraph{Genetically Evolved Majority CA} 
\cite{mitchellGA} studied how genetic algorithms can evolve CAs capable of global coordination. The authors were able to find a 1D CA denoted as $\phi_{par}$ with two states and radius $r=3$ which is successful at computing the majority task with the output required to be of the form of a homogenous state of either all 0's or all 1's. See Figure \ref{phi_par}.

This CA seems to belong to the Lin Class, which is confirmed by the linear regression fit with $R^2 \approx 99.2\% $.

\paragraph{Totalistic 1D 3-state CA} 
A totalistic CA is any CA whose local rule depends only on the number of cells in each state and not on their particular position. Wolfram studied various CA classes, one of them being the totalistic 1D CAs with radius $r=1$ and 3 states $S = \{0, 1, 2 \}$.

\begin{figure}[h!] 
\begin{tikzpicture}[thick, every node/.style={inner sep=0,outer sep=0}]
  \node at (4, +0.08) {
     \includegraphics[width=0.44\linewidth]{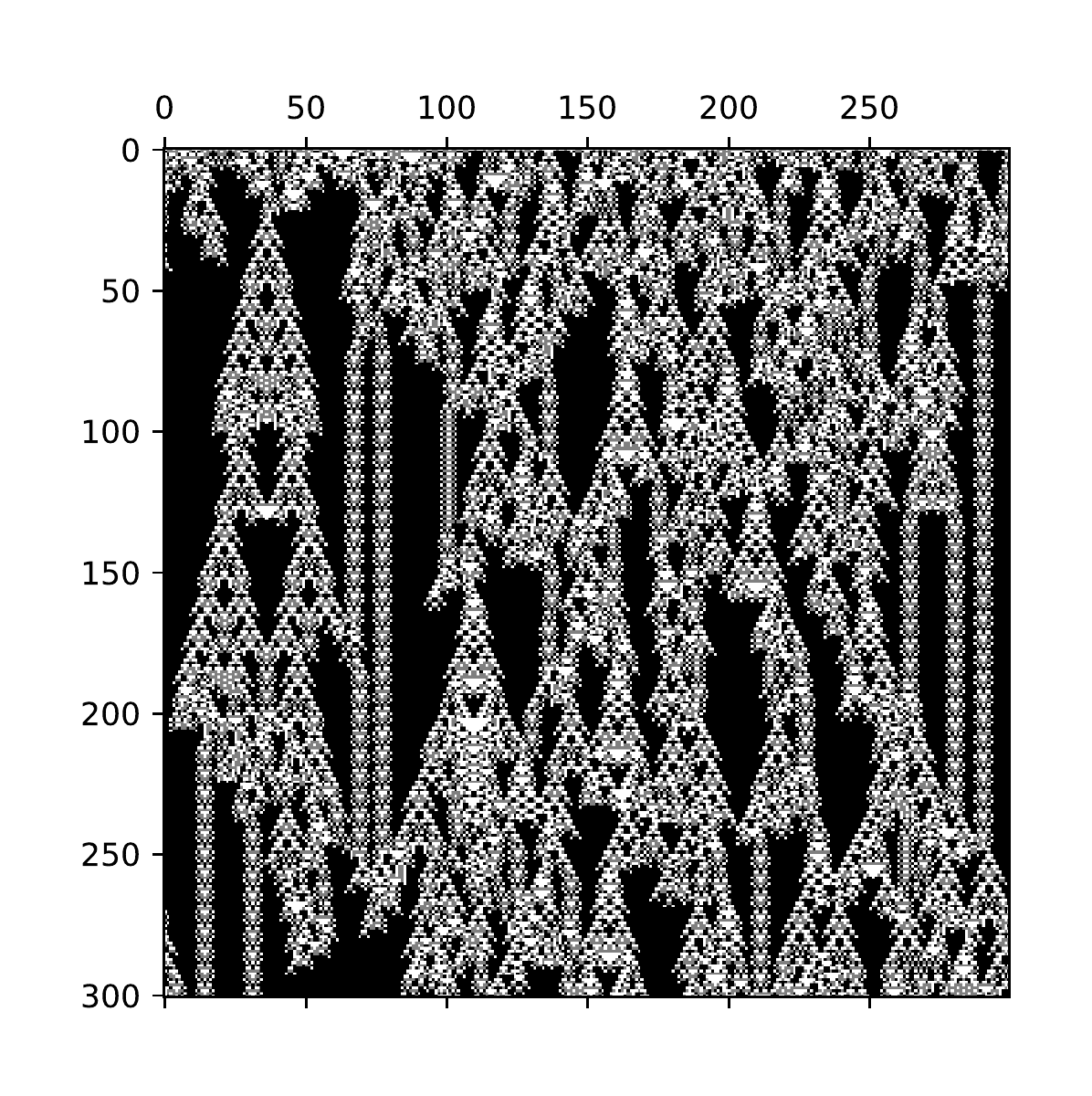}
  };
  \node at (-0.215, -0.1) {
     \includegraphics[width=0.6\linewidth]{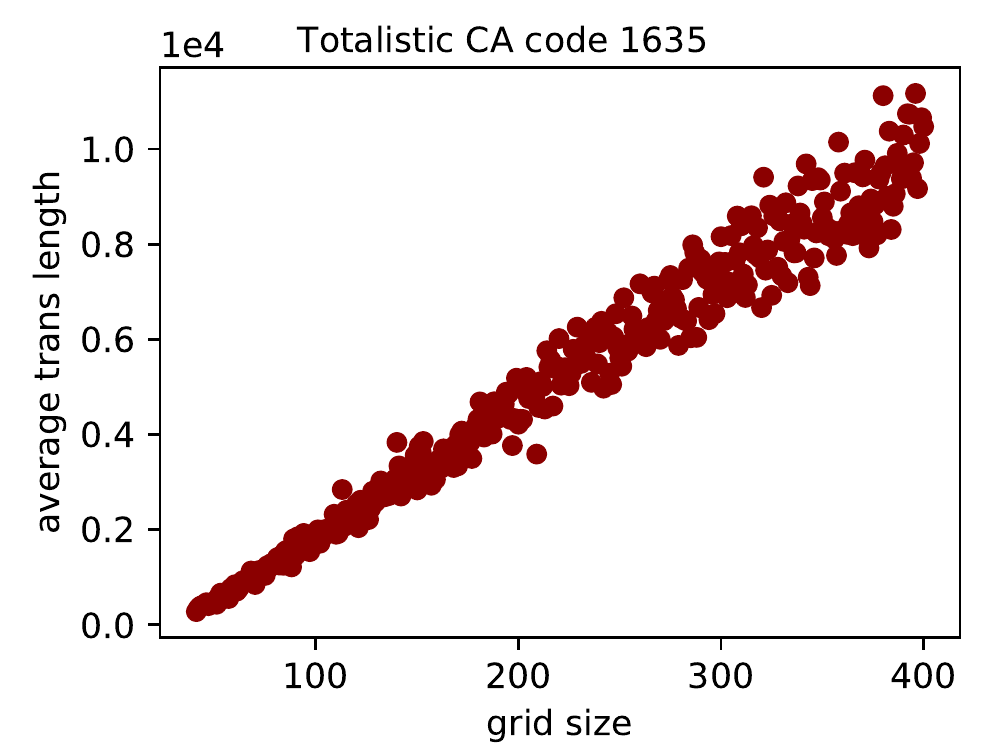}
  };

\end{tikzpicture}
\caption{Totalistic cellular automaton with code $1635$. The average transient growth plot is on the left. On the right, we show a space-time diagram of the evolution from a random initial configuration.}
\label{wolf_tot}
\end{figure}

\cite{newscience} presents a list of possibly complex CAs from this class. We applied the transient classification to such CA and learned that most of them were classified as logarithmic. This agrees with our space-time diagram observations that the local structures in such CAs ``die out'' quite quickly. Nonetheless, some of the CAs were classified as linear. An example of such a CA is in Figure \ref{wolf_tot} where the linear regression fit has $R^2 \approx 97.63\% $.

\section{Turing Machines}
In order to demonstrate the generality of the transient classification method, we further used it to examine the dynamics of Turing machines. In this section, we present the classification results.  

\subsection{Introducing Turing Machines}
Informally, a Turing machine (TM) consists of an infinite tape divided into cells and a movable reading head that scans one cell of the tape at a time. Every cell contains a symbol from some finite alphabet $A$, and the Turing machine is at an internal state from a finite set $S$. Depending on the symbol the head is reading and on its internal state, the Turing machine changes its internal state, rewrites the symbol on the tape, and either moves one cell to the left, right or stays in place. Turing machines represent the most classical model of computation; the Church-Turing thesis states that ``effectively calculable functions'' are exactly those that can be realized by a Turing Machine (\cite{turing36}). For a formal definition of Turing machines as well as a great introduction to computability theory, see \cite{soare}.

In this paper, we will consider deterministic Turing machines with one tape. $S$ will always denote the finite set of internal states, $A$ will denote the finite set of tape symbols. To ensure the Turing machine operates on a finite grid, as in the case of CAs, we will consider a tape of finite size with periodic boundary conditions. Therefore, each Turing machine with $S$ and $A$ operating on a tape of size $n$ gives rise to a global update function $$F: A^n \times \{1, 2, \ldots, n \} \times S \rightarrow  A^n \times \{1, 2, \ldots, n \} \times S,$$ where each configuration specifies the content of the tape, the position of the head, and the internal state. Thus, we can apply the transient classification to it.
We emphasize the non-traditional notion of the halting computation that we consider here. Classically, a Turing machine is considered to halt when it enters an attractor of size 1; that is when it does not change the tape’s content, the head’s position, or its internal state anymore. In our case, using the interpretation\begin{alignat*}{2}&\text{transients  }  &&\approx \text{  computation}\\&\text{attractors  }  &&\approx \text{  memory}\end{alignat*}we consider a Turing machine to halt whenever it enters \textbf{any} attractor. This is a much weaker notion of halting.

\begin{figure}[h!] 
\centering
\begin{tikzpicture}[thick, every node/.style={inner sep=0,outer sep=0}]
  \node at (0, +0.0) {
     \includegraphics[width=0.44\linewidth]{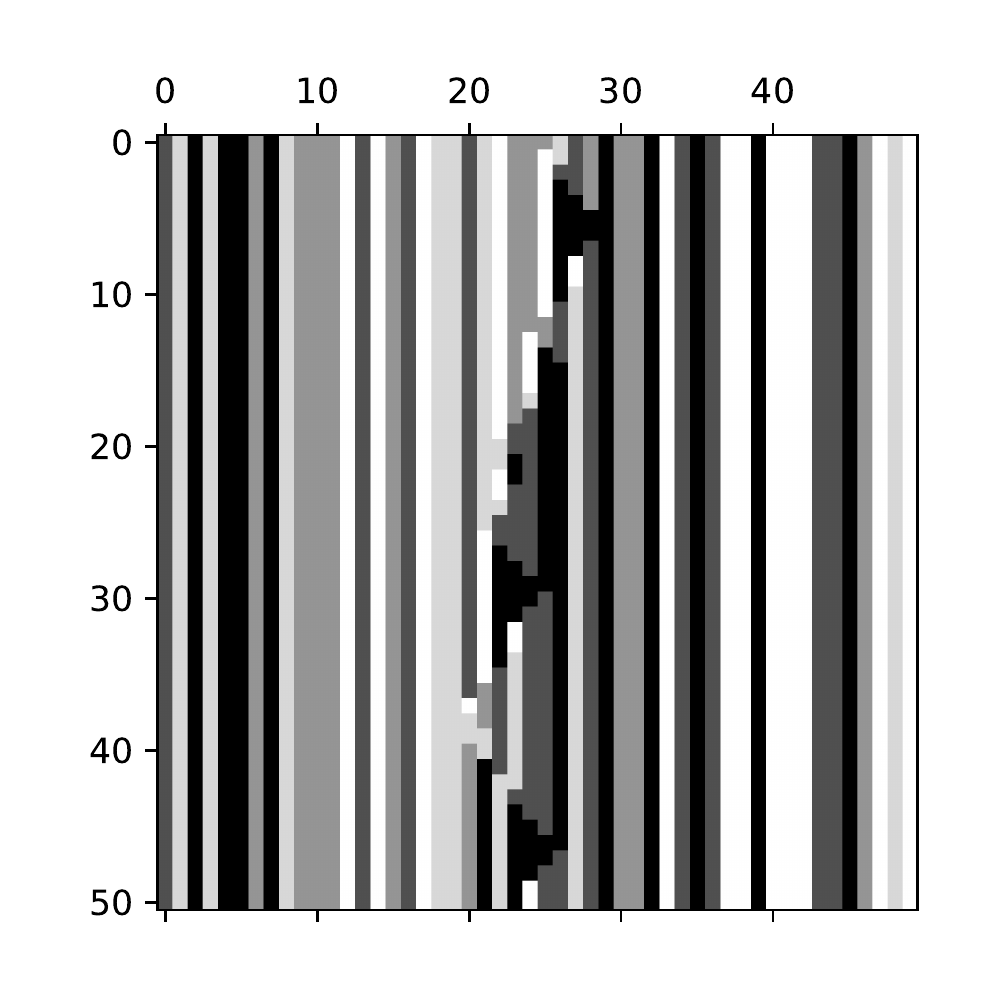}
  };
  \node at (3.5, 0) {
     \includegraphics[width=0.44\linewidth]{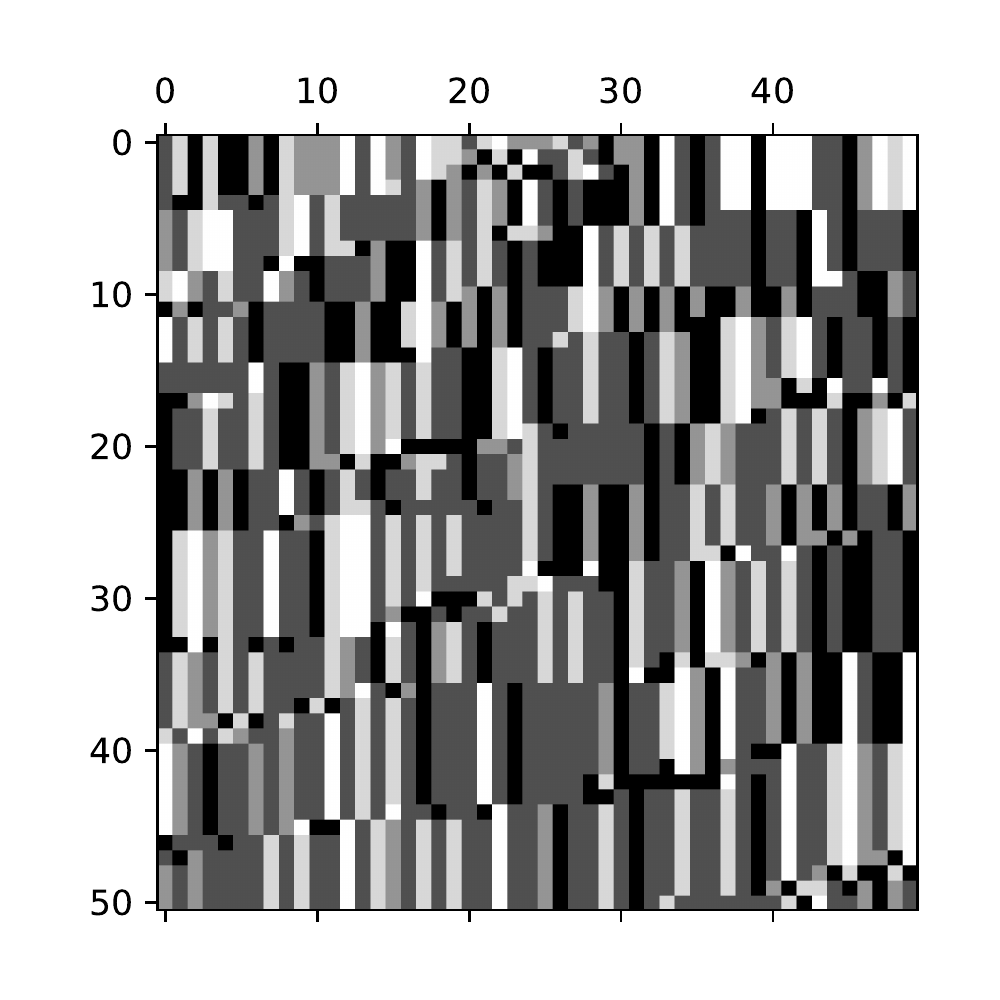}
  };

\end{tikzpicture}
\caption{Space-time diagrams of a 6 symbol, 5 state TM in the Exp Class. Classical space-time diagram is shown on the left. On the right we show the space-time diagram depicting the content of the tape after every 50 steps of computation on a tape of size 50.}
\label{sim_skip}
\end{figure}

We will depict the space-time diagrams of TM computation as a matrix, each row corresponds to the content of the tape at subsequent time steps, and time is progressing downwards. As opposed to CAs with their inherently parallel nature, TMs are sequential computational models. Thus, at each time step, only one symbol on the tape is changed. To produce space-time diagrams comparable to those produced by CAs, we only depict tape contents at every $n$-th step where $n$ is the size of the tape. This helps us to intuitively recognize the chaotic dynamics of TMs, an example is in Figure \ref{sim_skip}.

To the best of our knowledge, we know of no prior work examining the transients of Turing machines operating on cyclic tapes.

\subsection{Transient Classification of Turing Machines}
We have studied ``small'' TM with the number of states $|S|$ ranging from 4 to 8 and the number of alphabet symbols $|A|$ ranging from 2 to 5. For every such combination of values $|S|$ and $|A|$, we have randomly generated 100 transition functions of TMs and computed each of the TM's average transient length estimate for cyclic tapes of sizes ranging from 20 to 400. 

\subsubsection{Results}
For all the considered values of $|S|$ and $|A|$, more than $90 \%$ of the TMs were successfully classified using the transient classification method. We discuss a particular example in more detail below.

\paragraph{TMs with 7 states and 4 symbols} As an example, we present classification results of 100 Turing machines with 7 states and 4 tape symbols. The results are summarized in Table \ref{TM_7S_4A}.

\begin{table}[h!]
\centering
\begin{tabular}{ |m{3cm}||m{4cm}|  }
 \hline
 \multicolumn{2}{|c|}{Classification of TMs with 7 states and 4 symbols} \\
 \hline
 \centering
 Transient Class & \centering Percentage of TMs
 \tabularnewline
 \hline
Bounded Class & 41\%\\
Log Class &  2\%\\
Lin Class & 28\%\\
Poly Class & 13\%\\
Exp Class & 15\%\\
Unclassified & 1\%\\
 \hline
\end{tabular}
\caption{   Classification of 100 randomly sampled TMs with 7 states and 4 symbols.}
\label{TM_7S_4A}
\end{table}

\paragraph*{Bounded Class} (41/100 TMs).
In this space of rather ``small'' Turing machines, the Bounded Class seems to dominate the space. TMs in the Bounded Class halt in time independent on the tape size. Therefore, for such TMs it seems improbable to perform any nontrivial computation on both the finite and infinite tape.

\paragraph*{Log Class} (2/100 TMs)
The Log Class seems to be relatively small across all the TM classes we have examined. Here, the event that a TM head will read the whole input from the tape is improbable for large tape sizes.

\begin{figure}[h!] 
\begin{tikzpicture}[thick, every node/.style={inner sep=0,outer sep=0}]
  \node at (4, +0.08) {
     \includegraphics[width=0.44\linewidth]{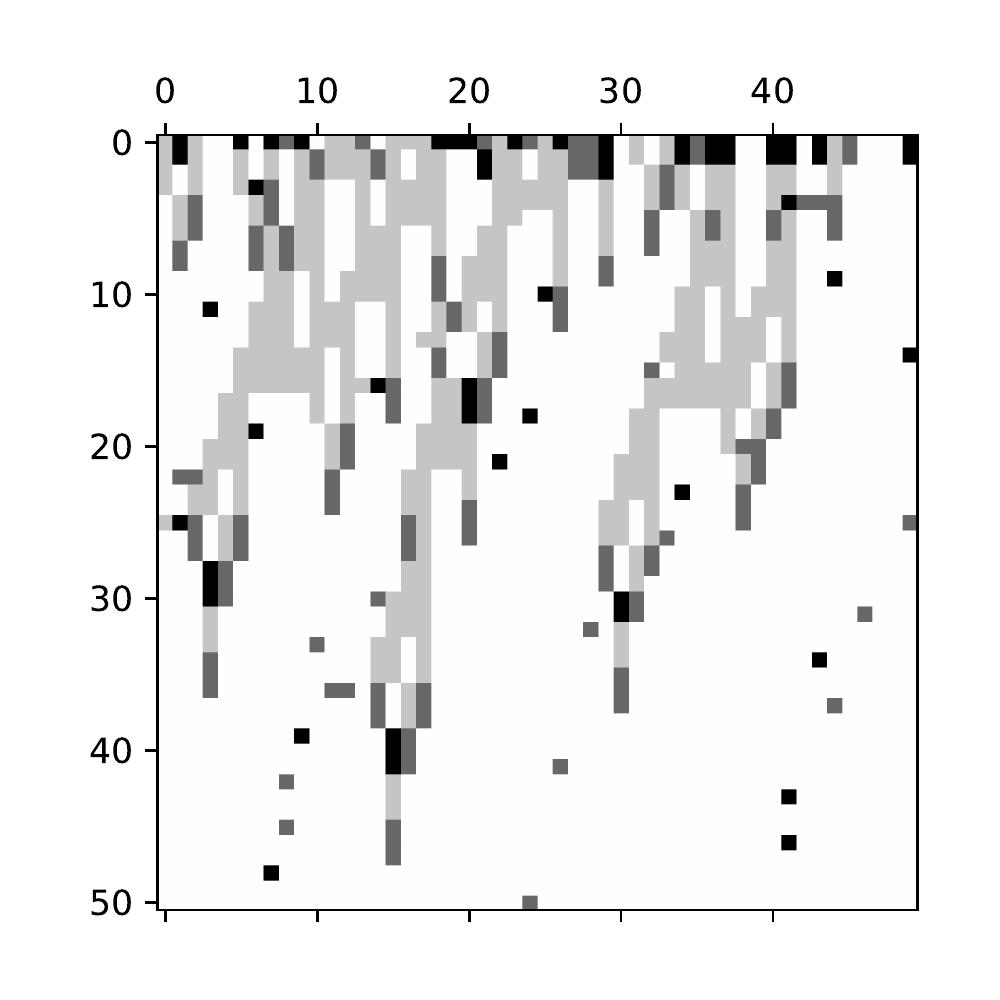}
  };
  \node at (-0.215, -0.1) {
     \includegraphics[width=0.6\linewidth]{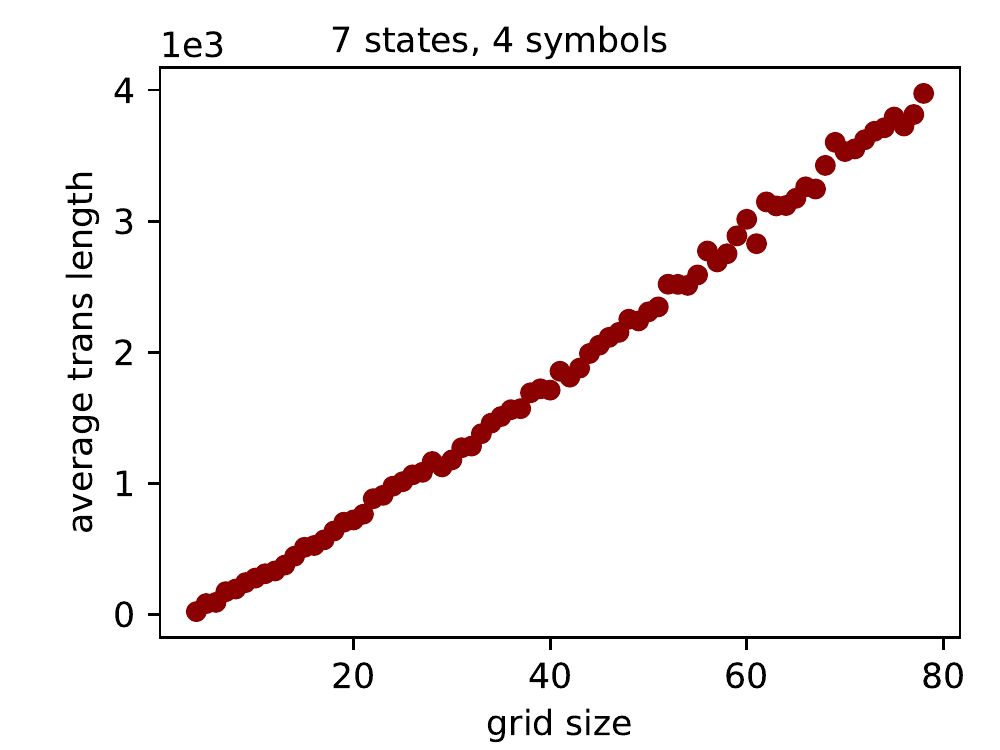}
  };
\end{tikzpicture}
\caption{Example of a Turing machine with 7 states, 4 symbols in the Lin Class. Its space-time diagram seems to exhibit nontrivial behavior.}
\label{7_4_lin}
\end{figure}

\paragraph*{Lin Class}(28/100 TMs.)
Let us consider a TM with trivial dynamics, which, given any input configuration, traverses each cell one by one and changes the state of each cell to the state $0 \in S$. After all cells enter this state, the computation halts. Such trivial behavior could be realized in constant time by a CA, though for a TM it takes at least $n$ steps where $n$ is the size of the tape. Hence, some Turing machines in the Lin Class exhibit periodic or simple dynamics. This emphasizes the fact that being contained in a Lin or Poly Class seems to be a necessary condition for complexity, not a sufficient one. See Figure \ref{7_4_lin}.

Nevertheless, we have observed TMs in the Lin Class whose space-time diagrams seem to contain some higher-level structures.

\paragraph*{Poly Class}(13/100 TMs.) In the Poly Class we have also observed TMs producing some higher-order structures.

\begin{figure}[h!] 
\begin{tikzpicture}[thick, every node/.style={inner sep=0,outer sep=0}]
  \node at (4, +0.08) {
     \includegraphics[width=0.44\linewidth]{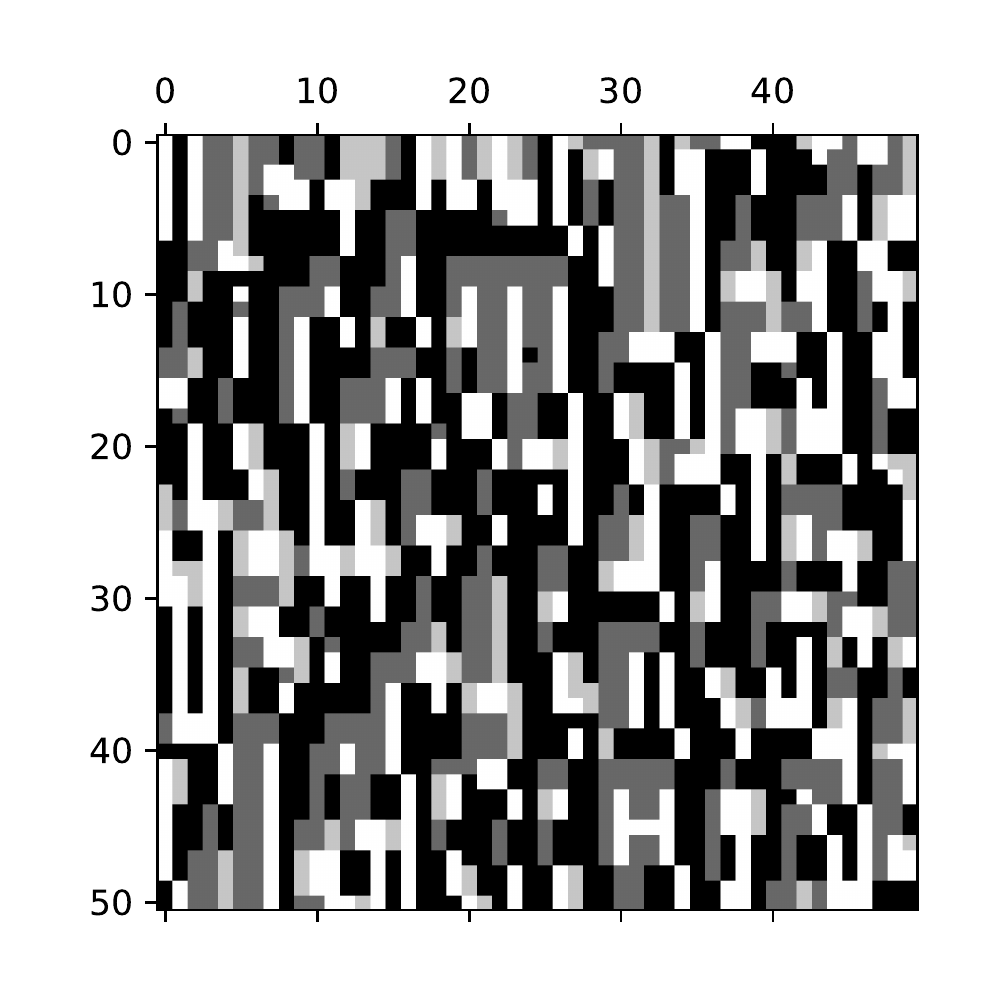}
  };
  \node at (-0.215, -0.1) {
     \includegraphics[width=0.6\linewidth]{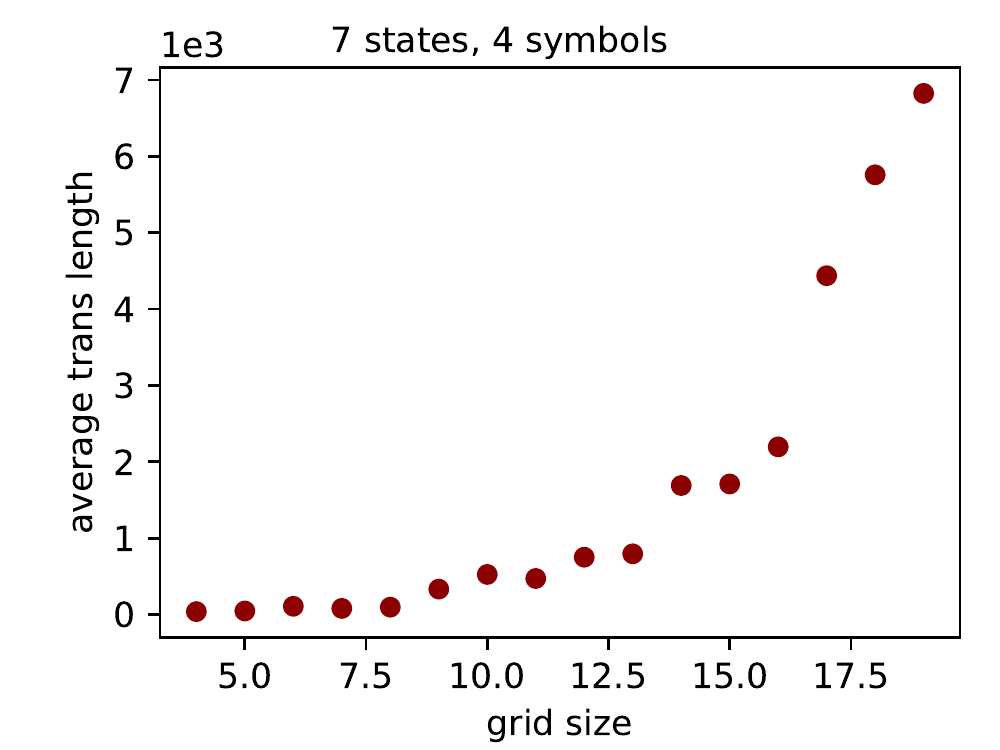}
  };
\end{tikzpicture}
\caption{Example of a Turing machine with 7 states, 4 symbols in the Exp Class.}
\label{7_4_exp}
\end{figure}

\paragraph*{Exp Class}(15/100 TMs.) We find it interesting that once only every $n$-th row of the space-time diagram ($n$ being the tape size) is depicted, the space-time diagrams of TMs in the Exp Class resemble the space-time diagrams of chaotic CAs. See Figure \ref{7_4_exp}.

As in the case of CAs, we are aware of the fact that the true asymptotic behavior of TMs in Lin or Poly Class might turn out to be logarithmic or exponential. In such a case, the systems in these classes would need significantly longer time to converge to their typical long-term behavior, which is a typical property of systems at a phase-transition.

\subsection{Transient Classification of Universal TMs}
\begin{figure}[h!] 
\begin{tikzpicture}[thick, every node/.style={inner sep=0,outer sep=0}]
 \node at (-0.215, -0.1) {
     \includegraphics[width=0.6\linewidth]{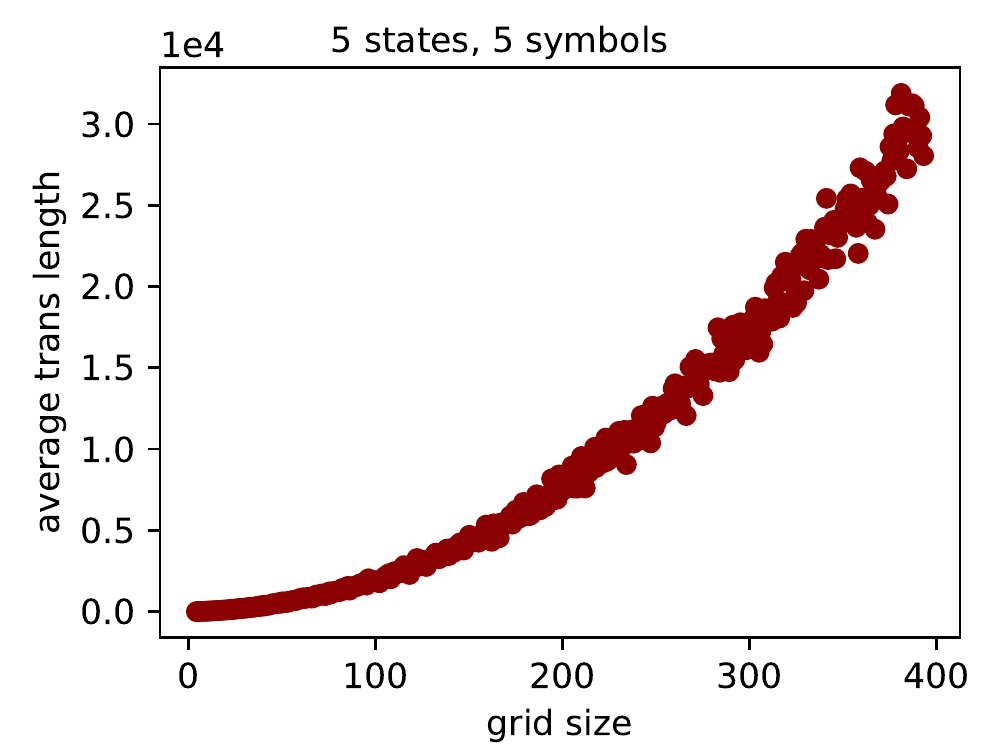}
  };
  \node at (4, +0.08)   {
     \includegraphics[width=0.44\linewidth]{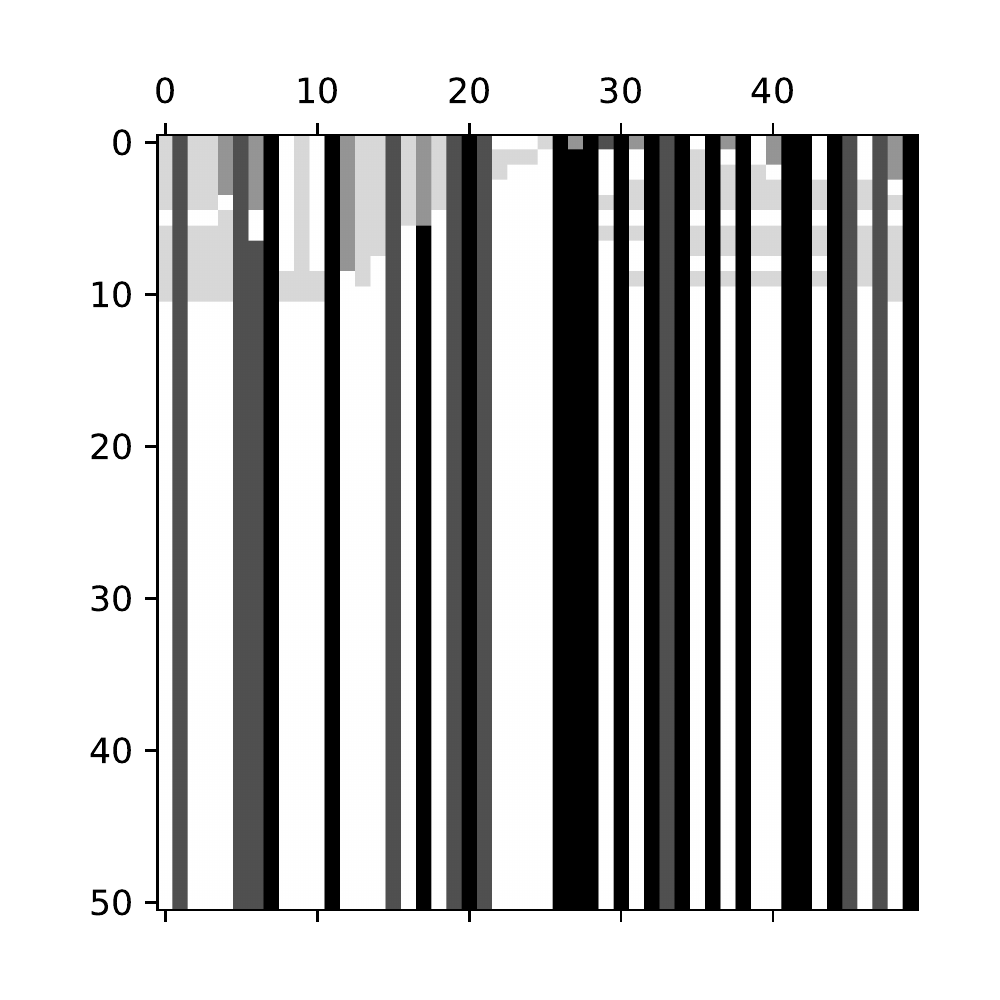}
  };
\end{tikzpicture}
\caption{Universal TMs with 10 states and 3 symbols belonging to the Poly Class. Again, every 50th step of the computation is shown in the space-time diagram.}
\label{universal_tm}
\end{figure}

Without much doubt, universal Turing machines are considered complex. We have estimated the asymptotic average computation time of 7 universal Turing machines with a small number of states and symbols constructed by \cite{rogozhin}. All of them were successfully classified, 6 belonging to the Poly class and 1 to the Lin Class. An example is shown in Figure \ref{universal_tm}.

Such results agree with the ones obtained for CAs, and support the hypothesis that complex dynamical systems belong to the Lin or Poly class.

\section{Random Boolean Networks}
Random Boolean networks form a very wide class of discrete dynamical systems that contains both CAs and TMs. In this section, we show that dynamical systems from this general class also conformed to our classification method.

\subsection{Introducing Random Boolean Networks}
The classical $N-K$ random Boolean network (RBN) is given by an oriented graph with $N$ nodes, each one of them having exactly $K$ edges pointing toward it. In addition, each node is equipped with a Boolean function of $K$ variables. Every node can have the value of either 0 or 1, therefore the configuration space is exactly $\{0, 1 \}^N$. To update a particular configuration of the network, the values of all nodes are changed in parallel, according to the outputs of their corresponding Boolean functions. This gives rise to a global update rule $F: \{0, 1 \}^N \rightarrow \{0, 1 \}^N$. For a concise introduction to RBNs see \cite{rbn_intro}.

RBNs were first introduced by Stuart \cite{kauff_rbn_intro} as models of gene regulatory networks. Classically, the nodes are interpreted as genes of a particular organism; their value represents whether the gene is ``turned on'' or ``off''. In this setting, the attractors of the network represent different cell types of the organism. Over the years, the networks have been widely studied as models of cell differentiation (\cite{huang_rbn_cell}), immune response (\cite{kauff_rbn_immune}), or neural networks (\cite{wuensche_rbn_nn}). The measures of criticality in RBNs were studied by \cite{luque_lyapunov}, \cite{fisher} or \cite{antifragility}. A great overview of work done on RBNs was written by \cite{rbn_overview}.

\subsubsection{Critical Behavior in RBNs}
RBNs are generic in the sense that both the connections of nodes and the Boolean functions are chosen uniformly at random. This makes it possible to analytically study the properties of a typical $N-K$ network. Indeed, different approaches (\cite{derrida_rbn}, \cite{luqsole_rbn}) lead to the same description of phase transitions in RBNs. We describe the results briefly below, as we will use them in our experiments.

We will describe a slightly more general model of RBNs. We consider a non-uniform connectivity of the nodes -- for a network of size $N$, we will assign to each node $i \in \{1, \ldots, N \}$ the connectivity $K_i \in \N$ and a Boolean function $f_i$ of arity $K_i$. Such a network is parametrized by the mean connectivity $\langle K  \rangle = \frac{1}{N} \sum_{i=1}^N K_i$. We also introduce the Boolean function sampling bias $p \in (0, 1)$. That is, we will sample the Boolean functions so that for all $i$ the probability that $f_i(x_1, \ldots x_{K_i})=1$ is $p$.
\cite{derrida_rbn} have analytically determined the edge of chaos for RBNs depending on the mean connectivity parameter $\langle K \rangle$ and Boolean function bias $p$. By studying the evolution of the distance between two randomly generated initial configurations over time, they have shown that the critical values of $\langle K \rangle$ and $p$ are exactly those satisfying
\begin{equation} \label{rbn_critical}
\langle{ K \rangle} = \frac{1}{2p (1-p)}.
\end{equation}
They obtain the following phases of RBN behavior.
\begin{alignat*}{2}
&\text{Ordered Phase } \ldots &&\text{ RBNs with } \langle{ K \rangle} < \frac{1}{2p (1-p)}\\
&\text{Critical Phase } \ldots &&\text{ RBNs with } \langle{ K \rangle} = \frac{1}{2p (1-p)}\\
&\text{Chaotic Phase } \ldots &&\text{ RBNs with } \langle{ K \rangle} > \frac{1}{2p (1-p)}\\
\end{alignat*}

\begin{figure}[h!] 
\centering
\includegraphics[width=0.6\linewidth]{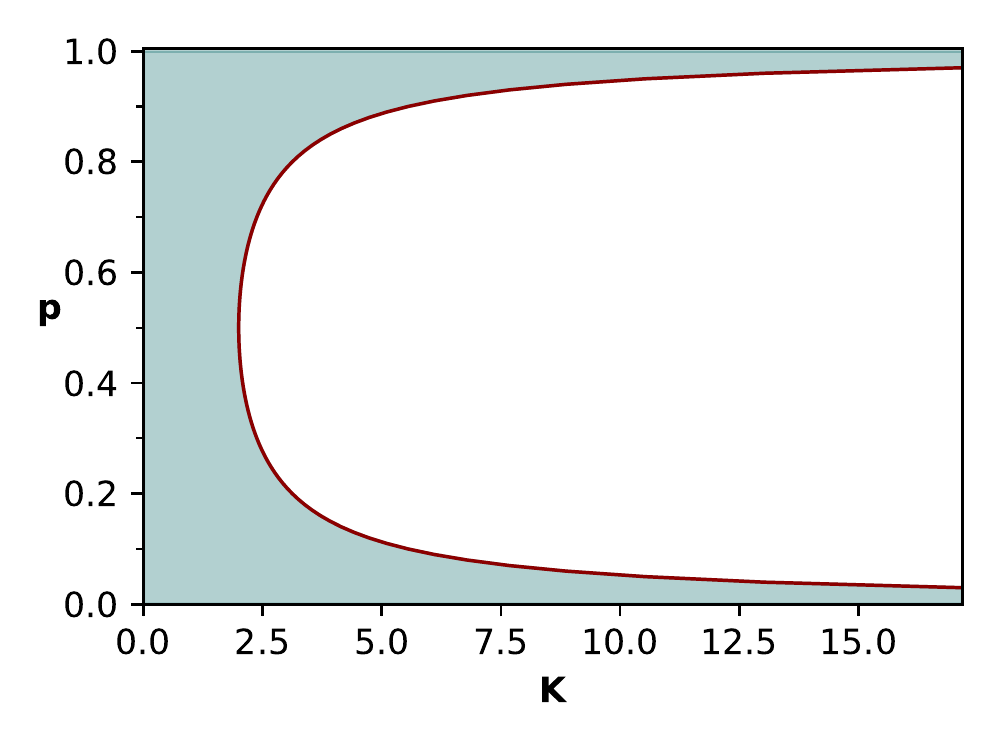}
\caption{Red curve depicts the critical values of RBN mean connectivity $\langle K \rangle$ and Boolean function bias $p$. The blue area denotes the region of ordered behavior, white area denotes the chaotic region.}
\label{rbn_critical_curve}
\end{figure}

The curve given by (1) is shown in Figure \ref{rbn_critical_curve}.

In the next section, we support the analytical results by showing that the transient classification clearly distinguishes between the ordered, critical, and chaotic regions.

\subsubsection{Phase-Space Properties of RBNs}
The generic nature of $N-K$ RBNs makes it possible to analytically study their global dynamics. This is a key difference between CAs and RBNs: CAs have a very particular architecture with only local connections and a uniform local transition rule. Therefore, the mean-field approximation methods of phase-space properties used for a ``typical'' $N-K$ RBN would not be as easy to apply to CAs.

With the classical interpretation of RBNs as gene regulatory networks where attractors represent different cell types, most of the focus has been on analyzing the number and size of the attractors for different values of $N$ and $K$. We briefly summarize some results related to our experiments below. For a more detailed discussion see \cite{rbn_overview}.

\paragraph{Ordered Phase}
In the ordered phase, when $K=1$, it has been shown that a probability of having an attractor of size $l$ falls exponentially with $l$ (\cite{flyv_rbnk1}). For a subset of $K=2$ RBNs with ordered behavior,  \cite{lynch_rbnk2} has shown that their average transient time grows at most logarithmically with the network size $N$.

\paragraph{Chaotic Phase}
In the case when $p=\frac{1}{2}$ and $K \geq N$, the RBN is essentially a random mapping whose phase-space properties have been studied extensively. It has been shown that both the average attractor and transient lengths of such RBNs grow exponentially with increasing $N$ (\cite{harris_rbnnk}, \cite{derrida_rbnnk}). 

We note that some previous work examining the transients of RBNs was conducted by \cite{rbn_trans1} and \cite{wuensche_rbn_nn} but we are not aware of any studies, which would use the asymptotic transient growth to describe the behavior of RBNs at the critical region.

\subsection{Transient Classification of RBNs}
We have sampled RBNs parametrized by the mean connectivity $\langle K \rangle$ and the Boolean function bias $p$. In this section we show that the results of transient classification clearly distinguish the ordered, critical, and chaotic phase of RBNs.

\subsubsection{Details of the Experiment}
Our goal is to estimate the average transient length of a ``typical'' RBN of size $N$, with mean connectivity $\langle K \rangle$ and Boolean function bias $p$. We would do so for increasing $N$ to observe the asymptotic behavior. We proceed as follows:

\begin{enumerate}
\item \label{step1} Given $N, \, \langle K \rangle$, and $p$, we generate a RBN $R(N,\langle K \rangle, p)$ with the corresponding parameters. We estimate the average transient length $T(R(N, \langle K \rangle, p))$ of $R(N, \langle K \rangle, p)$ using the approach described in Section \nameref{stats}.

\item We repeat step \ref{step1}. and generate a sequence of RBNs
$$R_1(N, \langle K \rangle, p), \, R_2(N, \langle K \rangle, p), \, \ldots, \,  R_m(N, \langle K \rangle, p)$$
and their average transient lengths 
$$T(R_1(N, \langle K \rangle, p)), \, \ldots, \, T(R_m(N, \langle K \rangle, p))$$
to ensure that we are close to the the average transient length of an average RBN with parameters $N$, $\langle K \rangle$, and $p$. We determine the number of sampled RBNs needed to get sufficiently close to the true average behavior by method analogous to the one described in \nameref{stats}.
Finally, we obtain the typical average transient length as
$$T(N, \langle K \rangle, p) = \frac{1}{m} \sum_{i=1}^m T(R_i(N, \langle K \rangle, p)).$$
\item We try to approximate the sequence $(T(N, \langle K \rangle, p))_{N=1}^\infty$ by generating a finite part of it. We typically compute $(T(N, \langle K \rangle, p))_{N=5}^{200}$, the upper bound being either $N=200$  or the limit imposed by the computation time of the transient lengths.
\end{enumerate}

\subsection{Results}

\begin{figure}[h!] 
\begin{tikzpicture}[thick, every node/.style={inner sep=0,outer sep=0}]
  \node at (0, 0) {
     \includegraphics[width=0.48\linewidth]{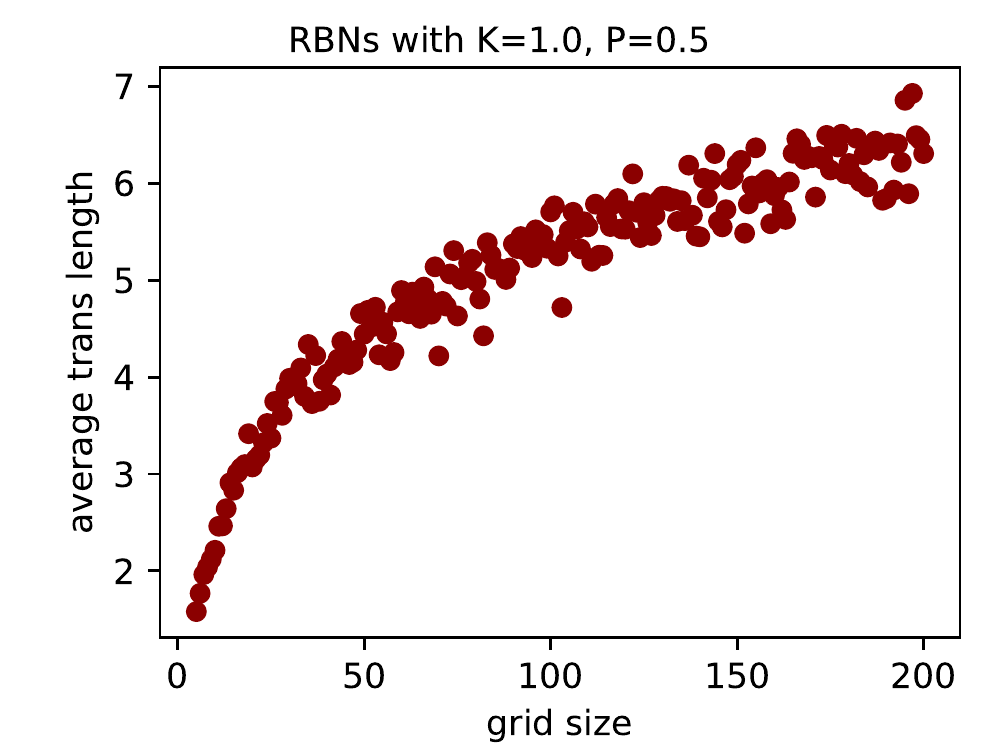}
  };
  \node at (4, 0) {
     \includegraphics[width=0.48\linewidth]{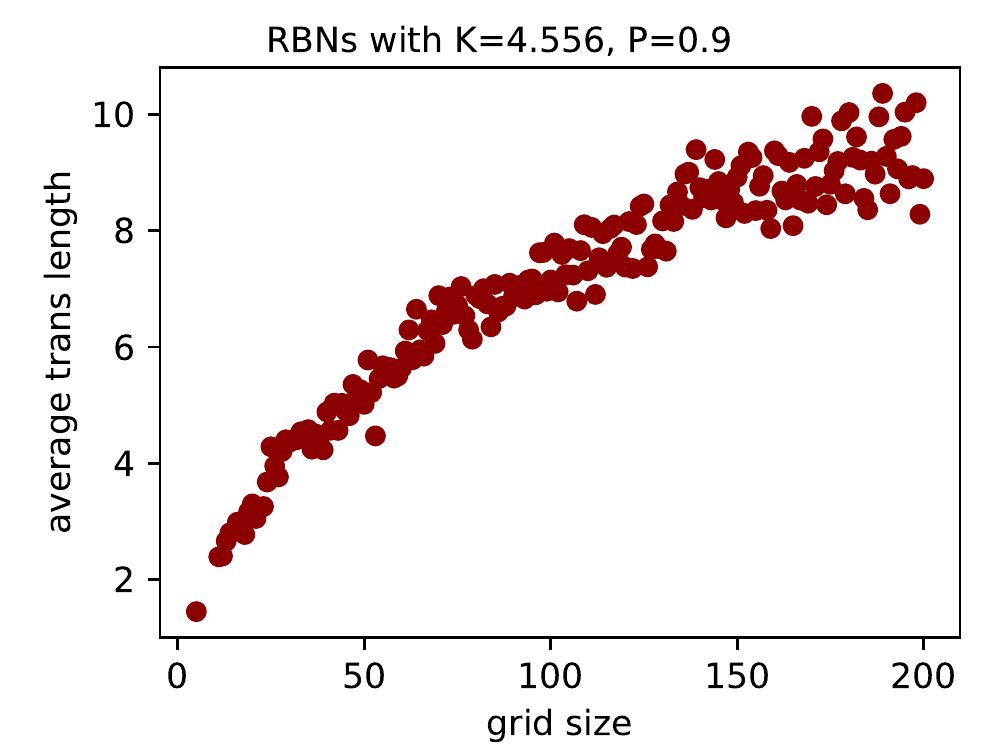}
  };

\end{tikzpicture}
\caption{Growth of typical average transient lengths for RBNs in the ordered region. RBN with mean connectivity $\langle K \rangle=1$ and Boolean function bias $p=0.5$ on the left, RBN with $\langle K \rangle=4.556$ and $p=0.9$ on the right. The best fit for both was logarithmic, with $R^2$ score over $90 \%$.}
\label{rbn_ordered}
\end{figure}

\paragraph{Ordered Phase} We have computed $ K_c, \, p$ along the curve given by (1) for $p=0.1, 0.2, \ldots, 0.9$ and sampled RBNs with parameters $\langle K_c-1 \rangle , \, p$ to ensure we are in the ordered region. See Figure \ref{rbn_ordered}.

For all such ensembles of RBNs the best fit for the typical average transient asymptotic growth was logarithmic. This supports the analytical results proven for special cases of $K$ and $p$ values.

\paragraph{Critical Phase} We have sampled RBNs with parameters $\langle K_c \rangle, \, p$ along the curve given by (1) for $p=0.1, 0.2, 0.3, \ldots, 0.9$. In all the sampled cases, the best fit for the typical average transient growth was linear. As in the case for CAs, we are aware that the asymptotic behavior of such RBNs can turn out to be logarithmic or exponential and that we just might not have sampled large enough networks. In such a case, we can interpret the Lin Class as a region of RBNs that take significantly longer to converge to their asymptotic behavior. See Figure \ref{rbn_critical}.

\begin{figure}[h!] 
\begin{tikzpicture}[thick, every node/.style={inner sep=0,outer sep=0}]
  \node at (0, 0) {
     \includegraphics[width=0.48\linewidth]{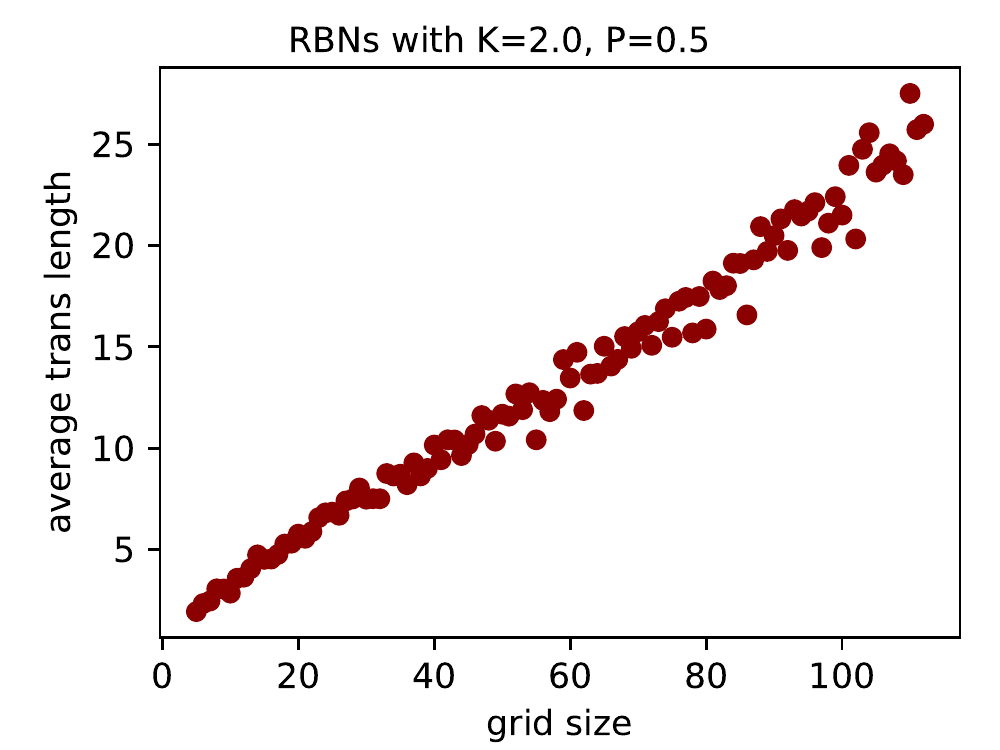}
  };
  \node at (4, 0) {
     \includegraphics[width=0.48\linewidth]{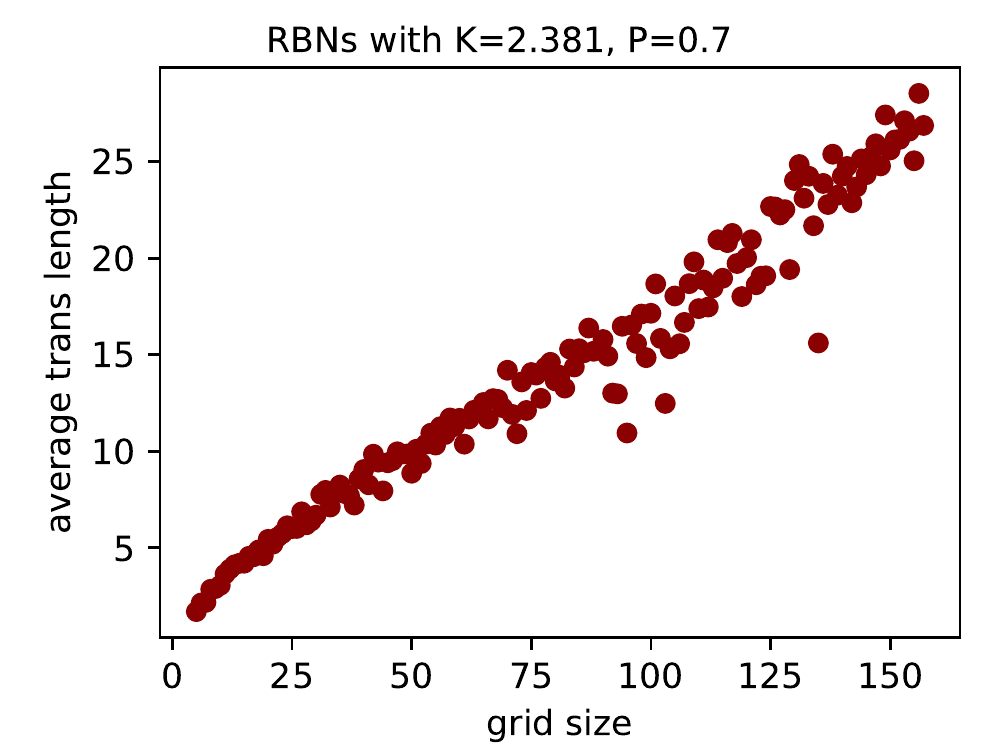}
  };

\end{tikzpicture}
\caption{Growth of typical average transient lengths for RBNs in the critical region. RBN with mean connectivity $\langle K \rangle=2$ and Boolean function bias $p=0.5$ on the left, RBN with $\langle K \rangle=2.381$ and $p=0.7$ on the right. The best fit for both was linear, with $R^2$ score over $95 \%$.}
\label{rbn_critical}
\end{figure}

\begin{figure}[h!] 
\begin{tikzpicture}[thick, every node/.style={inner sep=0,outer sep=0}]
  \node at (0, 0) {
     \includegraphics[width=0.48\linewidth]{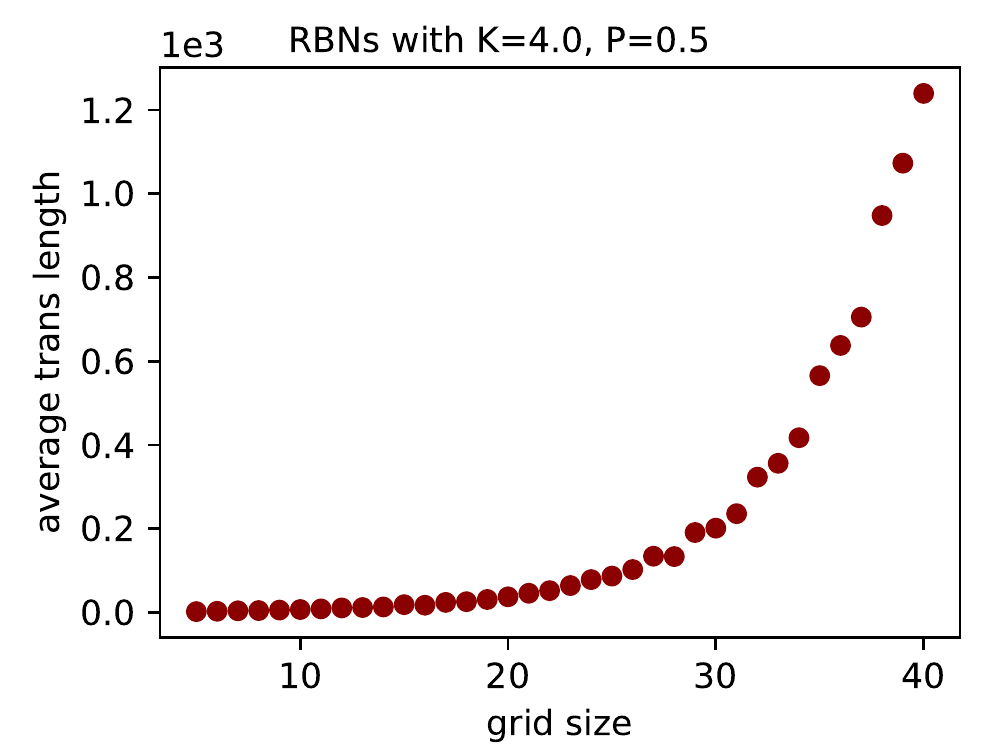}
  };
  \node at (4, 0) {
     \includegraphics[width=0.48\linewidth]{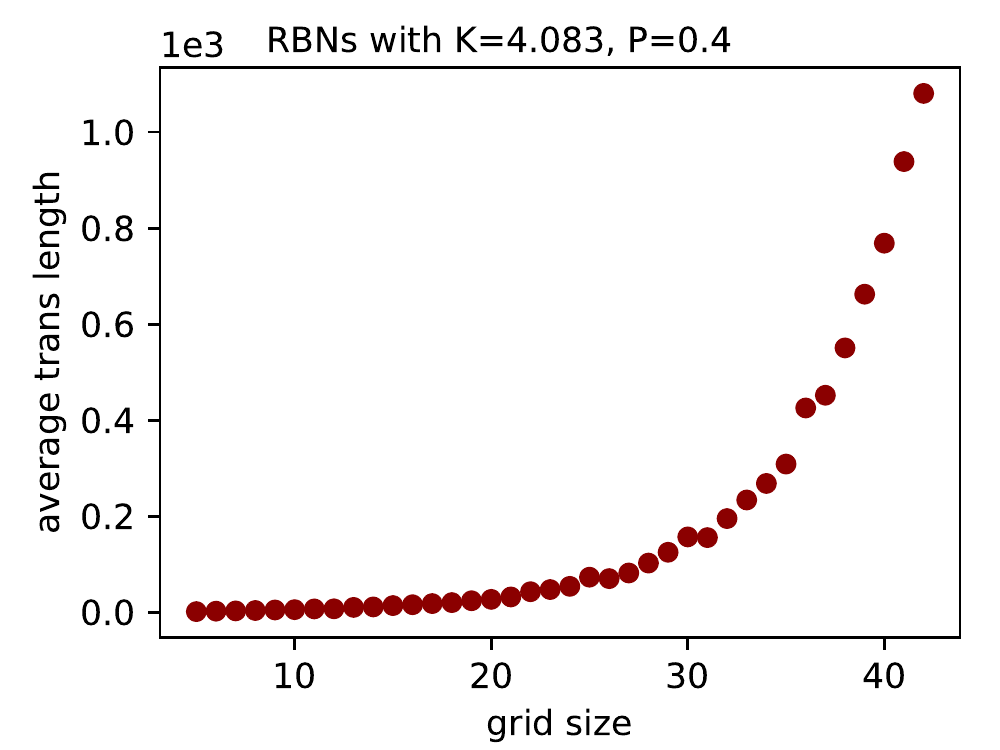}
  };

\end{tikzpicture}
\caption{Growth of typical average transient lengths for RBNs in the chaotic region. RBN with mean connectivity $K=4$ and Boolean function bias $p=0.5$ on the left, RBN with $K=4.083$ and $p=0.4$ on the right. The best fit for both was exponential, with $R^2$ score over $99 \%$.}
\label{rbn_chaotic}
\end{figure}

\paragraph{Chaotic Phase} We have computed the critical values $K_c, \, p$ along the curve given by (1) for values $p= 0.2, 0.3,  \ldots, 0.7, 0.8$ and sampled RBNs with parameters $\langle K_c+2 \rangle , \, p$ to ensure we are in the chaotic region. For all such ensembles of RBNs, the best fit for the typical average transient asymptotic growth was exponential, which again agrees with the analytic results. See Figure \ref{rbn_chaotic}.

These experiments support the results obtained for CAs and TMs indicating that ordered discrete systems belong to the Bounded or Log Class, chaotic systems correspond to the Exp Class, and complex systems lie in the region ``in between'', corresponding to the Lin and Poly Class.

\section{Conclusion}
We presented a classification method based on the asymptotic growth of average computation time. It is applicable to any deterministic discrete space and time dynamical system. We did present a good correspondence between the transient and Wolfram's classification in the case of ECAs. Further, we did show that the classification works for 2D CAs, Turing machines, and random Boolean networks, and we used it to discover 2D CAs capable of emergent phenomena. By demonstrating that complex discrete systems such as Game of Life, rule 110, several universal TMs, or RBNs with critical parameter values belong to the Lin or Poly Class, we believe that linear and polynomial transient growth navigates us toward a region of complex discrete systems.

Another elegant alternative would be to merge the Bounded and Log Class representing the ordered phase, and the Lin and Poly Class corresponding to the critical phase. In this way, we would obtain the traditional three phases of dynamics. This is entirely possible; we have kept the five classes to respect our initial experiments on elementary CAs where we obtained a much finer classification scheme.

The classification is based on a very simple idea and can be implemented with a few lines of code. In the field of ALife where novel discrete dynamical systems are designed as possible models of artificial evolution, the method we presented can be used to check whether such systems belong to the Lin or Poly Class. This might support the claim that such systems are capable of complex dynamics and emergent phenomena.

\section{Future Work}
We are interested in examining the transient growth of recurrent neural networks (RNNs). In the simplest case, we can add a ``rounding off'' output layer to discretize the configuration space. Then, the transient classification could be used to study the dynamics of RNNs, possibly guiding us towards appropriate network initializations and overall architectures yielding complex dynamics.

It would also be interesting to examine Busy Beaver Turing machines. Those are such TMs which take the longest time to halt (in the classical sense) among all TMs with the same number of states and tape symbols when run from an empty tape. It is interesting to observe how such machines behave when run from a randomly sampled initial configuration and whether they would exhibit complex dynamics, possibly being computationally universal (\cite{busy_beaver}).

Lastly, we could examine the dynamics of systems when simulated from a special region of its configuration space. The initial configurations could be generated by a designated algorithm, possibly discovering completely different dynamics of the system, as opposed to its average behavior.

\section{Acknowledgements}
Our work was supported by Grant Schemes at CU, reg.\ no.\ CZ.02.2.69/0.0/0.0/19\texttt{\char`_}073/0016935, the Ministry of Education, Youth and Sports within the dedicated program ERC CZ under the project POSTMAN with reference LL1902, by the Czech project AI$\&$Reasoning CZ.02.1.01/0.0/0.0/15\texttt{\char`_}003/0000466, European Regional Development Fund, by SVV-2020-260589, and is part of the RICAIP project that has received funding from the European Union's Horizon 2020 research and innovation programme under grant agreement No.\ 857306.

\footnotesize
\bibliographystyle{apalike}
\bibliography{trans_class} 

\end{document}